\documentclass[journal]{IEEEtran}

\usepackage{times}
\usepackage{epsfig}
\usepackage{graphicx}
\usepackage{amsmath}
\usepackage{amssymb}
\usepackage{subfigure}
\usepackage{multirow}
\usepackage{xcolor}
\usepackage[normalem]{ulem}
\usepackage{soul}
\useunder{\uline}{\ul}{}
\begin{document}

\title{Spatio-Temporal Saliency Networks\\ for Dynamic Saliency Prediction}
\author{Cagdas Bak, Aysun Kocak, 
	Erkut Erdem,
	and~Aykut Erdem
\thanks{The authors are with the Department
of Computer Engineering, Hacettepe University, Ankara, Turkey, TR-06800 (e-mail: cgds77@gmail.com; aysunkocak@cs.hacettepe.edu.tr; erkut@cs.hacettepe.edu.tr; aykut@cs.hacettepe.edu.tr)}}

\markboth{To appear in IEEE Transactions on Multimedia, 2017}%
{Shell \MakeLowercase{\textit{et al.}}: Bare Demo of IEEEtran.cls for IEEE Journals}

\maketitle

\begin{abstract}
Computational saliency models for still images have gained significant popularity in recent years. Saliency prediction from videos, on the other hand, has received relatively little interest from the community. Motivated by this, in this work, we study the use of deep learning for dynamic saliency prediction and propose the so-called spatio-temporal saliency networks. The key to our models is the architecture of two-stream networks where we investigate different fusion mechanisms to integrate spatial and temporal information. We evaluate our models on the DIEM and \mbox{UCF-Sports} datasets and present highly competitive results against the existing state-of-the-art models. We also carry out some experiments on a number of still images from the MIT300 dataset by exploiting the optical flow maps predicted from these images. Our results show that considering inherent motion information in this way can be helpful for static saliency estimation.
\end{abstract}


\IEEEpeerreviewmaketitle

\section{Introduction}
\label{sec:intro}

As a key part of the human visual system, visual attention mechanisms filter irrelevant visual stimuli in order to focus more on the important parts. Computational models of attention try to mimic this process through the use of machines and algorithms. These models have gained increasing attention lately. The reason behind this growing interest lies in their use in different computer vision and multimedia problems including but not limited to image retrieval~\cite{5730497}, visual quality assessment~\cite{7303954,7444164} video resizing/summarization~\cite{6529173,6527322}, action recognition~\cite{Wang2017}, event detection~\cite{Gan2015} either as a tool for visual feature extraction or as a mechanism for selecting features. These models are also important for generic applications such as advertisement and web design~\cite{7294708} as attention plays a key role in both user interfaces and human-machine interaction.

In the literature, the computational attention models developed so far generally aim to predict where humans fixate their eyes in images~\cite{borji2013state}. Specifically, they produce the so-called saliency maps from the visual data where a high saliency score at an image location indicates that the point is more likely to be fixated. These models are largely inspired by the hierarchical processing of human visual system~\cite{Hubel1962} and the theoretical studies like Feature Integration Theory~\cite{Treisman:CogPsych1980} or Guided Search Model~\cite{Wolfe:PsychBul1994}. They consider low-level image features (color, orientation, contrast, motion, etc.) and/or high-level features (pedestrians, faces, text) while predicting saliency maps. In this process, while low-level features are employed to examine how different an image point from its surroundings, high-level features are employed as it is experimentally shown that humans have a tendency to fixate on certain object classes more than others. 

Another current trend in the existing literature is to detect salient objects~\cite{6331539,borji2015salient,7347445,7514760} from the images. These models specifically aim at identifying the most prominent objects in an image that attract attention under free-viewing conditions and then segmenting them out from the background. These computational models for visual attention can be further grouped into two according to their inputs as static and dynamic saliency models. While static models work on images, dynamic models take video sequences as input. 

Saliency prediction from videos leads to great challenges when it is compared to carrying out the same task in still images. The reason is that the dynamic saliency frameworks require taking into account both spatial and temporal characteristics of the video sequences. Static saliency models use features like color, intensity and orientation, however dynamic models need to focus more on the moving objects or image parts as it is shown that humans there is a tendency for humans to look at them while viewing. Hence, the preliminary models proposed for dynamic saliency prediction extend the existing saliency models suggested for still images so that they consider extra motion features~\cite{harel2006graph,guo2008spatio,cui2009temporal,seo2009static}. However, more recent works approach the same task from a different point of view and propose novel solutions~\cite{hou2007saliency,mathe2012dynamic,rudoy2013learning}.

\subsection{Overview of our approach}
Deep learning has been successfully applied to saliency prediction in still images in the last few years, providing state-of-the-art results~\cite{Kümmerer2014b,pan2016shallow,kruthiventi2015deepfix,vig2014large,zhao2015saliency,Bruce_2016_CVPR,Jetley_2016_CVPR}. The early models utilize pre-trained deep convolutional neural networks (CNNs) proposed to classify images as generic feature extractors and build classifiers on top of those features to classify fixated image regions~\cite{vig2014large,Kümmerer2014b}. Later models, however, approach the problem from an end-to-end perspective and either train networks from scratch or most of the time fine-tune the weights of a pre-trained model~\cite{Kruthiventi_2016_CVPR, Jetley_2016_CVPR,pan2016shallow}. The modifications in the network architectures are usually about integrating multi-scale processing or using different loss functions~\cite{HuangSALICON2015,Jetley_2016_CVPR}. It has been investigated that the power of these deep models mainly comes from the property that the features learned by these networks are semantically very rich~\cite{Bruce_2016_CVPR}, capturing high-level factors important for saliency detection. Motivated by the success of these works, in this study, we explore the use of two-stream CNNs for saliency prediction from videos. To the best of our knowledge, our work is the first deep model for dynamic saliency, which is trained in an end-to-end manner, that learns to combine spatial and temporal information in an optimal manner within a two-stream network architecture. 

\subsection{Our contributions}
The contributions of our work can be summarized as follows:
\begin{enumerate}
\item  We study two-stream convolutional neural networks which mimic the visual pathways in the brain and combine networks trained on temporal and spatial information to predict saliency map of a given video frame. Although these network architectures have been previously investigated for some computer vision problems such as video classification~\cite{karpathy2014large} and action recognition~\cite{simonyan14b}, to our knowledge, we are the first to apply two-stream deep models for saliency prediction from videos in the literature. In particular, in our study, we investigate two different fusion strategies, namely element-wise and convolutional fusion strategies, to integrate spatial and temporal streams.

\item We carry out extensive experiments on DIEM ~\cite{mital2011clustering} and UCF-Sports~\cite{MatheSminchisescuPAMI2015} datasets and compare our deep spatio-temporal saliency networks against several state-of-the-art dynamic saliency models. Our  evaluation demonstrates that the proposed STSConvNet model outperforms these models in nearly all of the evaluation metrics on these datasets. 

\item On a number of challenging still images, we also show that our spatio-temporal saliency network can predict the human fixations better than the state-of-the-art deep static saliency models. The key idea that we follow is to extract optical flow from these static images by using a recently proposed method~\cite{walker2015dense} and feed them to our network along with the appearance image. 
 
\end{enumerate}

\section{Related Work} 
In this study, we focus on bottom-up modeling of dynamic saliency. Below, we first summarize the existing dynamic saliency models from the literature and then provide a brief overview of the proposed deep-learning based static saliency models which are related to ours.

\subsection{Dynamic Saliency} 
Early examples of saliency models for dynamic scenes extend the previously proposed static saliency models which process images in a hierarchical manner by additionally considering features related to motion such as optical flow. For instance, in~\cite{harel2006graph}, Harel \emph{et al.} propose a graph-theoretic solution to dynamic saliency by representing the extracted feature maps in terms of fully connected graphs and by predicting the final saliency map. In~\cite{cui2009temporal}, Cui \emph{et al.} extract salient parts of video frames by performing spectral residual analysis on the Fourier spectrum of these frames over the spatial and the temporal domains. In~\cite{guo2008spatio}, Guo \emph{et al.} propose a similar spectral analysis based formulation. In~\cite{sultani2014human}, Sultani and Saleemi extend Harel \emph{et al.}~\cite{harel2006graph}'s model by using additional features such as color and motion gradients and by post-processing the predicted maps via a graphical model based on Markov Random Fields. In~\cite{seo2009static}, Seo and Milanfar employ self similarities of spatio-temporal volumes to predict saliency. ~\cite{mauthner2015encoding}, Mauthner \emph{et al.} also present a video saliency detection method for using as a prior information for activity recognition algorithms. Instead of using a data driven approach they propose an unsupervised algorithm for estimating salient regions of video sequences.   

Following these early works, other researchers rather take different perspectives and devise novel solutions for dynamic saliency. For example, in~\cite{hou2007saliency}, Hou and Zhang consider rarity of visual features while extracting saliency maps from videos and propose an entropy maximization-based model. In~\cite{rahtu2010segmenting}, Rahtu \emph{et al.} extract saliency by estimating local contrast between feature distributions. In~\cite{Ren2012}, Ren \emph{et al.} propose a unified model with the temporal saliency being estimated by sparse and low-rank decomposition and the spatial saliency map being extracted by considering local-global contrast information. In~\cite{mathe2012dynamic}, Mathe \emph{et al.} devise saliency prediction from videos as a classification task where they integrate several visual cues through learning-based fusion strategies. In another study, Rudoy \emph{et al.}~\cite{rudoy2013learning} propose another learning based model for dynamic saliency. It differs from Mathe \emph{et al.}'s model~\cite{mathe2012dynamic} in that they take into account a sparse set of gaze locations thorough which they predict conditional gaze transitions over subsequent video frames. Zhou \emph{et al.}~\cite{zhou2014learning} oversegment video frames and use low-level features from the extracted regions to estimate regional contrast values. Zhao \emph{et al.}~\cite{zhao2015fixation} learn a bank of filters for fixations and use it to model saliency in a location-dependent manner. Khatoonabadi \emph{et al.}~\cite{hossein2015many} propose a saliency model that depends on compressibility principle. In a very recent study, Leboran \emph{et al.}~\cite{awsd}, propose another dynamic saliency model by using the idea that perceptual relevant information is carried by high-order statistical structures.

\subsection{Deep Static Saliency} 
In recent years, deep neural networks based models provide state-of-the-art results in many computer vision problems such as image classification~\cite{he2015deep}, object detection~\cite{girshick2015fast}, activity recognition~\cite{zhou2014learning}, semantic segmentation~\cite{long2015fully} and video classification~\cite{karpathy2014large}. These approaches perform hierarchical feature learning specific to a task, and thus gives results better than the engineered features. Motivated by the success of these models, a number of researchers have recently proposed deep learning based models for saliency prediction from images~\cite{Kümmerer2014b,pan2016shallow,kruthiventi2015deepfix,vig2014large,Bruce_2016_CVPR,Jetley_2016_CVPR,Li2016}. 

Vig \emph{et al.}~\cite{vig2014large} use an ensemble of CNNs which learns biologically inspired hierarchical features for saliency prediction. K\"{u}mmerer \emph{et al.}~\cite{Kümmerer2014b} employ deep features learned through different layers of the AlexNet~\cite{krizhevsky2012imagenet} and learn how to integrate them for predicting saliency maps. Kruthiventi et al.~\cite{kruthiventi2015deepfix} adopt VGGNet~\cite{simonyan15b} for saliency estimation where they introduce a location-biased convolutional layer to model the center-bias, and train the model on SALICON dataset using Euclidean loss. Jetley \emph{et al.}~\cite{Jetley_2016_CVPR} also use the VGGNet architecture but they especially concentrate on investigating different kinds of probability distance measures to define the loss function. Pan \emph{et al.}~\cite{pan2016shallow} very recently propose two CNN models having different layer sizes by approaching saliency prediction as a regression task. Li \emph{et al.}~\cite{Li2016} employ a fully convolutional neural network within a multi-task learning framework to jointly detect saliency and perform object class segmentation. It is important to note that all these models are proposed for predicting saliency in still images not videos. Bruce \emph{et al.}~\cite{Bruce_2016_CVPR} propose yet another fully convolutional network to predict saliency and they try to understand factors and learned representations when training these type of networks for saliency estimation.

Motivated by the deep static saliency models, in our paper we investigate the use of two-stream CNNs for saliency prediction from videos. In fact, investigating layered formulations is not new for saliency prediction. As discussed earlier, most of the traditional dynamic saliency models are inspired from the hierarchical processing in the low-level human vision~\cite{Hubel1962}. These models, however, employ hand-crafted features to encode appearance and motion contrast to predict where humans look at in dynamic scenes. Since they depend on low-level cues, they often fail to capture semantics of scenes at its full extent, which is evidently important for gaze prediction. More recent models, on the other hand, employ learning-based formulations to integrate these low-level features with the detection maps for faces, persons, and other objects. This additional supervision boosts the prediction accuracies, however, the performance is limited by the discrimination capability of the considered features and the robustness of the employed detectors.

As compared to the previous dynamic saliency models, our deep spatio-temporal saliency networks are trained to predict saliency in an end-to-end manner. This allows us to learn hierarchical features, both low-, mid- and high-level,~\cite{Cichy2016,bylinskii2016should} that are specialized for the gaze prediction task. For instance, while the early layers learn filters that are sensitive to edges or feature contrasts, the filters in the top layers are responsible from capturing more complex, semantic patterns that are important for the task. In our case, our deep two-stream saliency networks learn multiple layers of spatial and temporal representations and ways to combine them to predict saliency.

In particular, in our study we extract temporal information via optical flow between consecutive video frames and investigate different ways to use this additional information in saliency prediction within a deep two-stream spatio-temporal network architecture~\cite{simonyan14b}. These two-stream networks are simple to implement and train, and to our interest, are in line with the hierarchical organization of the human visual system. Specifically, the biological motivation behind these architectures is the so-called two-streams hypothesis~\cite{two-stream-hypothesis} which speculate that human visual cortex is comprised of two distinct streams, namely ventral and the dorsal streams, which are respectively specialized to process appearance and motion information. 

Here, an alternative deep architecture could be to stack two or more frames together and feeding this input to a deep single-stream CNN, which was investigated in several action recognition networks~\cite{7410867,6165309,Taylor2010,7410879}. In this work, we do not pursue this direction because of two reasons. Firstly, this approach requires learning 3D convolutional filters~\cite{7410867,6165309,Taylor2010} in order to capture spatio-temporal regularities among input video frames but using 3D filters highly increases the complexity of the models and these 3D convolutional networks are harder to train with limited training data~\cite{7410879} (which is the case for the existing dynamic saliency datasets). Secondly, 3D convolutional filters are mainly used for expressing long-range motion patterns which could be important for recognizing an action since they cannot easily be captured by optical-flow based two-stream models. For dynamic saliency prediction though, we believe that such long-range dependencies are minimally important as human attention shifts continuously, and optical flow information is sufficient to establish the link between motion and saliency.

\begin{figure*}[!t]
\centering
\includegraphics[width=\textwidth]{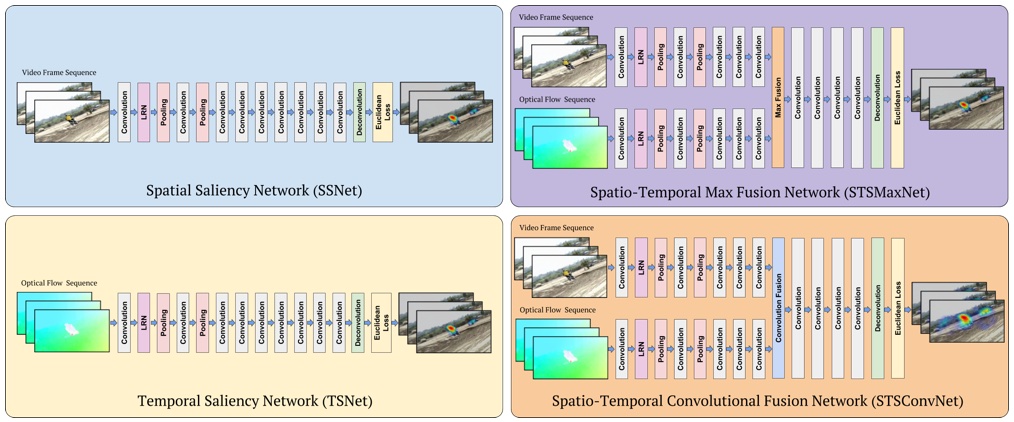}	\\
\begin{tabular}{cp{1cm}p{1cm}c}
\small{(a) Single stream saliency networks} & & & \small{(b) Two-stream saliency networks}\\	
\end{tabular}
\caption{(a) The baseline single stream saliency networks. While SSNet utilizes only spatial (appearance) information and accepts still video frames, TSNet exploits only temporal information whose input is given in the form of optical flow images. (b) The proposed two-stream spatio-temporal saliency networks. STSMaxNet performs fusion by using element-wise max fusion, whereas STSConvNet employs convolutional fusion after the fifth convolution layers.}
\label{fig:saliency-networks}
\end{figure*}

\section{Our Models}
The aim of our study is to investigate the use of deep architectures for predicting  saliency from dynamic scenes. Recently, CNNs provided drastically superior performance in many classification and regression tasks in computer vision. While the lower layers of these networks respond to primitive image features such as edges, corners and shared common patterns, the higher layers extract semantic information like object parts, faces or text~\cite{Bruce_2016_CVPR,bylinskii2016should}. As mentioned before, such low and high-level features are shown to be both important and complementary in estimating visual saliency. Towards this end, we examine two baseline single stream networks (spatial and temporal) given in Figure~\ref{fig:saliency-networks}(a) and two two-stream networks~\cite{simonyan14b} shown in Figure~\ref{fig:saliency-networks}(b), which combine spatial and temporal cues via two different integration mechanisms: element-wise max fusion and convolutional fusion, respectively. We describe these models in detail below.

\subsection{Spatial Saliency Network}
For the basic single stream baseline model, we retrain the recently proposed static saliency model in~\cite{pan2016shallow} for dynamic saliency prediction by simply ignoring temporal information and using the input video frame alone. Hence, this model does not consider the inter-frame relationships while predicting saliency for a given video. As shown in the top row of Figure~\ref{fig:saliency-networks}(a), this CNN resembles the VGG-M model~\cite{simonyan15b} -- the main difference being that the final layer is a deconvolution (fractionally strided convolution) layer to up sample to the original image size. Note that it does not use any temporal information and exploits only appearance information to predict saliency in still video frames. We refer to this network architecture as SSNet.

\subsection{Temporal Saliency Network}
Saliency prediction from videos is inherently different than estimating saliency from still images in that our attention is highly affected by the local motion contrast of the foreground objects. To understand the contribution of temporal information to the saliency prediction, we develop a second single stream baseline. As given in the bottom row of Figure~\ref{fig:saliency-networks}(a), this model is just a replica of the spatial stream net but the input is provided in the form of optical flow images, as in~\cite{simonyan14b}, computed from two subsequent frames. We refer to this single stream network architecture as TSNet.

\subsection{Spatio-Temporal Saliency Network with Direct Averaging}
As a baseline model, we define a network model which integrates the responses of the final layers of the spatial and the temporal saliency networks by using direct averaging. Note that this model does not consider a learning strategy on how to combine these two-stream network and consider each one of the single-stream networks equally reliable. We refer to this two-stream network architecture as STSAvgNet.

\subsection{Spatio-Temporal Saliency Network with Max Fusion} 
This network model accepts both a video frame and the corresponding optical flow image as inputs and merges together the spatial and temporal single stream networks via element-wise max fusion. That is, given two feature maps $\mathbf{x}^{s},\mathbf{x}^{t}\in \mathbb{R}^{H\times W \times D}$ from the spatial and temporal streams, with $W,H,D$ denoting the width, height and the number of channels (filters), max fusion takes the maximum of these two feature maps at every spatial location $i$ and $j$, and channel $d$, as: 
\begin{equation}
	y^{max}_{i,j,d} = \max\left(x^{s}_{i,j,d} , x^{t}_{i,j,d}\right)\;.
\end{equation}

The use of the $\max$ operation makes the ordering of the channels in a convolutional layer arbitrary. Hence, this fusion strategy assumes arbitrary correspondences between the spatial and temporal streams. That said, this spatio-temporal model seeks filters so that these arbitrary correspondences between feature maps become as meaningful as possible according to the joint loss. After this fusion step, it also uses a deconvolution layer to produce an up-sampled saliency map as the final result as illustrated in the top row of Figure~\ref{fig:saliency-networks}(b). We refer to this two-stream network architecture as STSMaxNet.

\subsection{Spatio-Temporal Saliency with Convolutional Fusion} 
This network model integrates spatial and temporal streams by applying convolutional fusion. That is, the corresponding feature maps $\mathbf{x}^{s}$ and $\mathbf{x}^{t}$ respectively from the spatial and temporal streams are stacked together and then combined as follows:
\begin{equation}
	\mathbf{y}^{conv} = \left[ \mathbf{x}^{s} \quad\mathbf{x}^{t}\right]*\mathbf{f}+\mathbf{b}\;,
\end{equation}
where $\mathbf{f}\in\mathbb{R}^{1\times 1 \times 2D \times D}$ denotes a bank of $1\times 1$ filters, and $\mathbf{b}\in \mathbb{R}^D$ represents the bias term.

The main advantage of the convolutional fusion over the element-wise max fusion is that the filterbank $\mathbf{f}$ learns the optimal correspondences between the spatial and temporal streams based on the loss function, and reduces the number of channels by a factor of two through the weighted combinations of $\mathbf{x}^{s}$ and $\mathbf{x}^{t}$ with weights given by $\mathbf{f}$ at each spatial location. As demonstrated in the bottom row of Figure~\ref{fig:saliency-networks}(b), this is followed by a number of convolutions and a final deconvolution layer to produce the saliency map. We refer to this network architecture as STSConvNet.

\subsection{Spatio-Temporal Saliency Network with Direct Fusion}
Finally, as another baseline model, we design a single stream network in which the appearance and optical flow images are stacked together and fed to the network as input. This model implements an early fusion strategy at the very beginning of the network architecture and can be seen as a special case of STSConvNet. Here, each layer of the network  learns a set of filters that directly acts on the given appearance and motion frames. We refer to this model as STSDirectNet.

\section{Implementation Details}
\label{ssec:implementation_details}

\subsection{Network Architectures}
The architecture of our single stream models is the same with that of the deep convolution network proposed in~\cite{pan2016shallow}. They take $320\times 240\times 3$ pixels images and processes them by the following operations: $C(96,7,3)$ $\rightarrow$ $LRN$ $\rightarrow$ $P$ $\rightarrow$ $C(256,5,2)$ $\rightarrow$ $P$ $\rightarrow$ $C(512,3,1)$ $\rightarrow$ $C(512,5,2)$ $\rightarrow$ $C(512,5,2)$ $\rightarrow$ $C(256,7,3)$ $\rightarrow$ $C(128,11,5)$ $\rightarrow$ $C(32,11,5)$ $\rightarrow$ $C(1,13,6)$ $\rightarrow$ $D$. Here, $C(d,f,p)$ represents a convolutional layer with $d$ filters of size $f\times f$ applied to the input with padding $p$ and stride 1. $LRN$ denotes a local response normalization layer that carries out a kind of lateral inhibition, and $P$ indicates a max pooling layer over $3\times 3$ regions with stride 2. Finally, $D$ is a deconvolution layer with filters of size $8\times 8\times 1$ with stride 4 and padding 2 which upscales the final convolution results to the original size. All convolutional layers except the last one are followed by a ReLU layer. Our spatial and  temporal stream models in particular differ from each other in their inputs. While the first one processes still images, the next one accepts optical flow images as input.

For the proposed spatio-temporal saliency networks shown in Figure~\ref{fig:saliency-networks}(b), we employ element-wise max and convolutional fusion strategies to integrate the spatial and temporal streams. Performing fusion after the fifth convolutional layer gives the best results for both of these fusion strategies. In STSMaxNet, the single stream networks are combined by applying element-wise max operation, which is followed by the same deconvolution layer in the single stream models. On the other hand, STSConvNet performs fusion by stacking the feature maps together and integrating them by a convolution layer $C(512,1,0)$ whose weights are initialized with identity matrices. The remaining layers are the same with those of the single stream models.

\subsection{Data Preprocessing} 
\begin{figure}[!t]
	\centering
	\begin{tabular}{c@{\;}c@{\;}c}
		\includegraphics[width=0.15\textwidth,height=1.8cm]{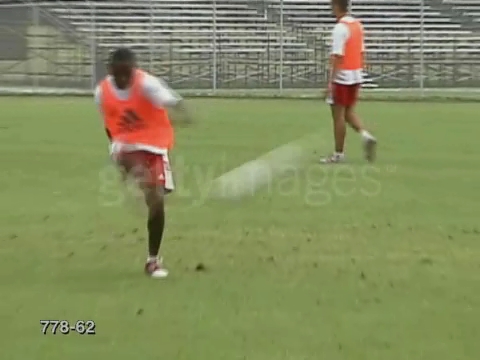} &
		\includegraphics[width=0.15\textwidth,height=1.8cm]{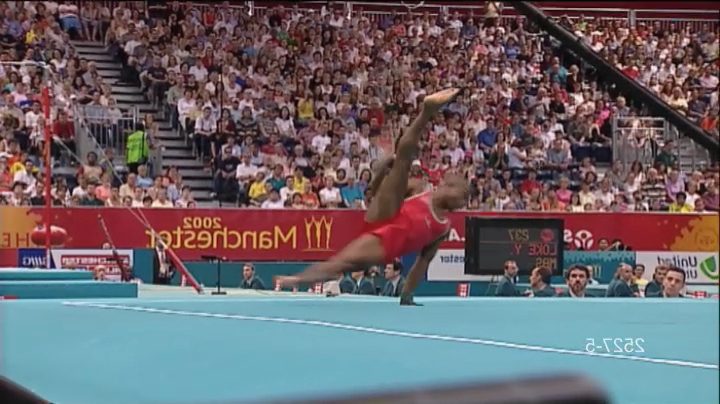} &
		\includegraphics[width=0.15\textwidth,height=1.8cm]{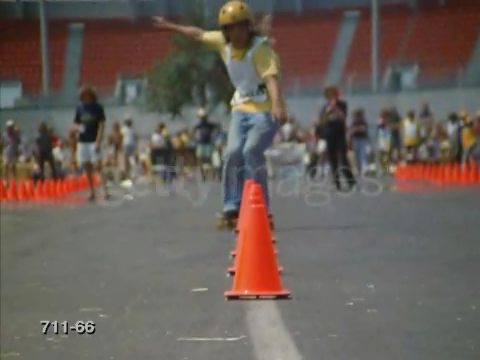}\\
		\includegraphics[width=0.15\textwidth,height=1.8cm]{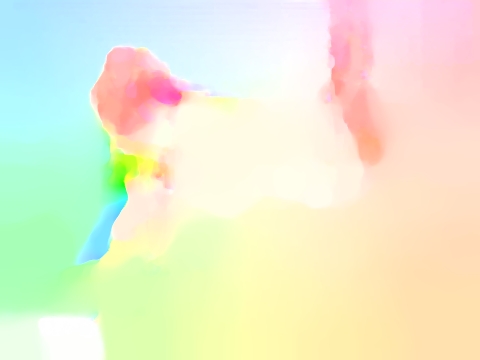} &
		\includegraphics[width=0.15\textwidth,height=1.8cm]{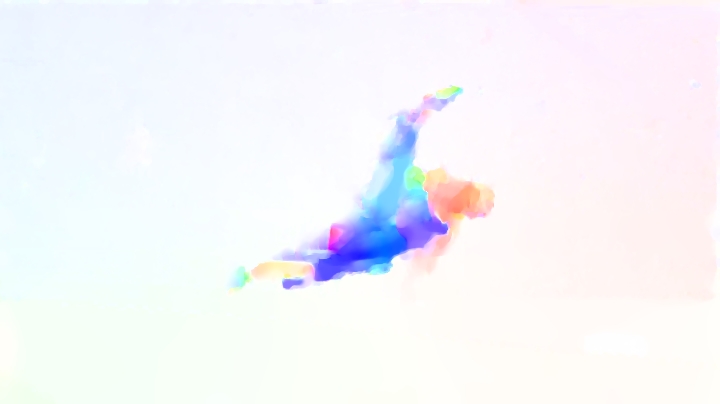} &
		\includegraphics[width=0.15\textwidth,height=1.8cm]{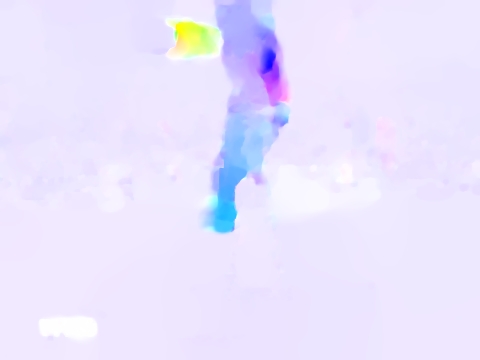}
	\end{tabular}
	\caption{Sample optical flow images generated for some frames of a video sequence from UCF-Sports dataset.\vspace{-6pt}}
	\label{fig:opticalflow}
\end{figure}

We employ three publicly available datasets, 1.DIEM (Dynamic Images and Eye Movements)~\cite{mital2011clustering}, 2. UCF-Sports~\cite{MatheSminchisescuPAMI2015} datasets and 3. MIT 300 dataset~\cite{mit-saliency-benchmark}, which are described in detail in Section~\ref{sec:experiments}, in our experiments. Since our networks accept inputs of size $320\times 240\times 3$ pixels and outputs saliency maps of the same size, all videos and ground truth fixation density maps are rescaled to this size prior to training. We use the publicly available implementation of DeepFlow~\cite{weinzaepfel2013deepflow} and we additionally extract optical flow information from the rescaled versions of subsequent video frames. Optical flow images are then generated by stacking horizontal and vertical flow components and the magnitude of the flow together. Some example optical flow images are shown in Figure~\ref{fig:opticalflow}. 

\subsection{Data Augmentation}
Data augmentation is a widely used approach to reduce the effect of over-fitting and improve generalization of neural networks. For saliency prediction, however, classical techniques such as cropping, horizontal flipping, or RGB jittering are not very suitable since they alter the visual stimuli used in the eye tracking experiments in collecting the fixation data. Having said that, horizontal flipping is used in~\cite{pan2016shallow} as a data augmentation strategy although there is no theoretical basis for why this helps to obtain better performance.

In our study, we propose to employ a new and empirically grounded data augmentation strategy for specifically training saliency networks. In~\cite{judd2011fixations}, the authors performed a thorough analysis on how image resolution affects the exploratory behavior of humans through an eye-tracking experiment. Their experiments revealed that humans are quite consistent about where they look on high and low-resolution versions of the same images. Motivated with this observation, we process all video sequences and produce their low-resolution versions by down-sampling them by a factor of 2 and 4, and use these additional images with the fixations obtained from original high-resolution images in training. We note that in reducing the resolution of optical flow images the magnitude should also be rescaled to match with the down-sampling rate. It is worth-mentioning that this new data augmentation strategy can also be used for boosting performances of deep models for static saliency estimation.

\subsection{Training} 
We employ the weights of the pretrained CNN model in~\cite{pan2016shallow} to set the initial weights of our spatial and temporal stream networks. In training the models, we use Caffe framework~\cite{jia14} and employed Stochastic Gradient Descent with Euclidean distance between the predicted saliency map and the ground truth. The networks were trained over 200K iterations where we used a batch size of 2 images, momentum of 0.9 and weight decay of 0.0005, which is reduced by a factor of 0.1 at every 10K iterations. Depending on the network architectures, it takes between 1 day to 3 days to train our models on the DIEM dataset by using a single 2GB GDDR5 NVIDIA GeForce GTX 775M GPU on a desktop PC equipped with 4-core Intel i5 (3.4 GHz) Processor and 16 GB memory. At test time, it takes approximately 2 secs to extract the saliency map of a single video frame.

\section{Experimental Results}
\label{sec:experiments}
In the following, we first review the datasets on which we perform our experiments and provide the list of state-of-the-art computational saliency models that we compared against our spatio-temporal saliency networks. We then provide the details of the quantitative evaluation metrics that are used to assess the model performances. Next, we discuss our experimental results. 

\subsection{Datasets}
To validate the effectiveness of the proposed deep dynamic saliency networks, our first set of experiments is carried out on the DIEM dataset~\cite{mital2011clustering}. This dataset is collected for the purpose of investigating where people look at dynamic scenes. It includes 84 high-definition natural videos including movie trailers, advertisements, etc. Each video sequence has eye fixation data collected from approximately 50 different human subjects. In our experiments, we only used the right-eye positions of the human subjects as done in~\cite{borji2013quantitative}.

We perform our second set of experiments on the UCF-Sports ~\cite{MatheSminchisescuPAMI2015}. This dataset is collected from broadcast television channels such as the BBC and ESPN which consists of a set of sport actions~\cite{MatheSminchisescuPAMI2015}. The video sequences are collected from wide range of websites. This dataset contains 150 video sequences with $720\times 480$ resolution and cover a range of scene and viewpoints. The dataset includes several actions such as diving, golf swing, kicking and lifting, and is used for action recognition. However, recently, additional human gaze annotations were collected in~\cite{MatheSminchisescuPAMI2015}. These fixations are collected over 16 human subjects under task specific and task-independent free viewing conditions.

Lastly, for the experiments on predicting eye fixations on still images, we choose a number of images from the MIT 300 dataset~\cite{mit-saliency-benchmark}. Selected images are the ones especially depicting an action and including objects that are interpreted as in motion. This dataset has eye fixation data collected from 39 subjects under free-viewing conditions for 3 secs for a total of 300 natural images with longest dimension 1024 pixels and the other dimension varied from 457 to 1024 pixels.  

\subsection{The compared computational saliency models}
Through our experiments on DIEM and UCF-Sports datasets, we compare our deep network models with eight state-of-the-art dynamic saliency models: GVBS~\cite{harel2006graph}, PQFT~\cite{guo2008spatio}, SR~\cite{hou2007saliency}, Seo and Milanfar~\cite{seo2009static}, Rudoy \emph{et al.}~\cite{rudoy2013learning}, Fang \emph{et al.}~\cite{fang2014}, Zhou \emph{et al.}~\cite{zhou2014learning}, and DWS~\cite{awsd} models. Moreover, we compare our two-stream deep models STSMaxNet and STSConvNet to a certain extent with deep static saliency model DeepSal~\cite{pan2016shallow} as the architectures of our TSNet and SSNet models are adapted from this model.

\subsection{Evaluation Measures}
We employ Area Under Curve (AUC), shuffled AUC (sAUC)~\cite{zhang2008sAUC}, Pearson's Correlation Coefficient (CC), Normalized Scanpath Saliency (NSS)~\cite{nss}, Normalized Cross Correlation (NCC) and ${\chi}^2$ distance throughout our experiments for performance evaluation. We note that NCC and ${\chi}^2$ distance are not widely-used measures but we report them as some recent studies~\cite{rudoy2013learning,zhao2015fixation,mauthner2015encoding} employ them in their analysis.

AUC measure considers the saliency maps as a classification map and uses the receiver operator characteristics curve to estimate the effectiveness of the predicted saliency maps in capturing the ground truth eye fixations. An AUC value of~1 indicates a perfect match with the ground truth while the performance of chance is indicated by a value close to 0.5. The AUC measure cannot account for the tendency of human subjects to look at the image  center of the screen, i.e. the so-called center bias. Hence, we also report the results of the shuffled version of AUC (sAUC) where the center bias is compensated by selecting the set of fixations used as the false positives from  another randomly selected video frame from the dataset instead of the processed frame.

CC treats the predicted saliency and the ground truth human fixation density maps as random variables and measures the strength of a linear relationship between two using a Gaussian kernel density estimator. While a value close to 0 indicates no correlation, a CC value close to +1/-1 demonstrates a good linear relationship. NSS estimates the average normalized saliency score value by examining the responses at the human fixated locations on the predicted saliency map which has been normalized to have zero mean and unit standard deviation. While a NSS value of 0 indicates chance in predicting eye fixtions, a non-negative NSS value, especially that of greater than 1, denotes correspondence between maps above chance.

NCC is a general measure used for assessing image similarity. It treats the ground truth fixation map and the predicted saliency map as images and estimates a  score with values close to 1 implying high similarity and negative values indicating low similarity. ${\chi}^2$ distance considers the saliency maps as a probability distribution map and compares the predicted map with the ground truth human fixation map accordingly. A perfect prediction model needs to provide a distance close to~0 for the ${\chi}^2$ distance. 

\begin{figure*}[!t]
	\centering
	\begin{tabular}{c@{\;}c@{\;}c@{\;}c@{\;}c@{\;}c@{\;}c}
		\includegraphics[width=0.135\textwidth]{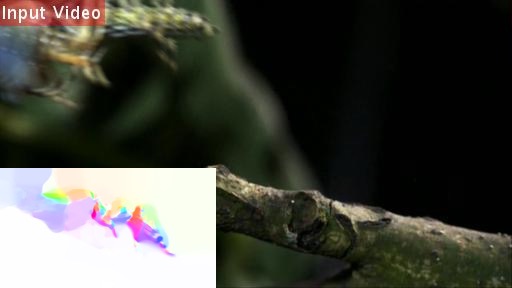} &
		\includegraphics[width=0.135\textwidth]{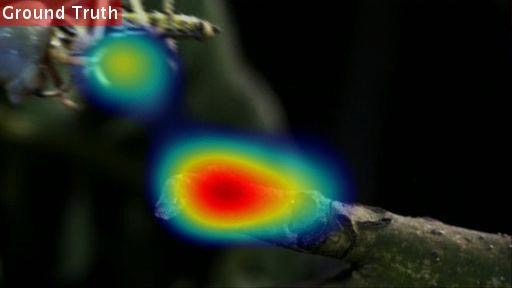} &	
		\includegraphics[width=0.135\textwidth]{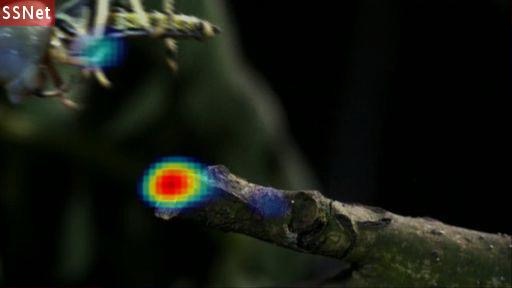} &	
		\includegraphics[width=0.135\textwidth]{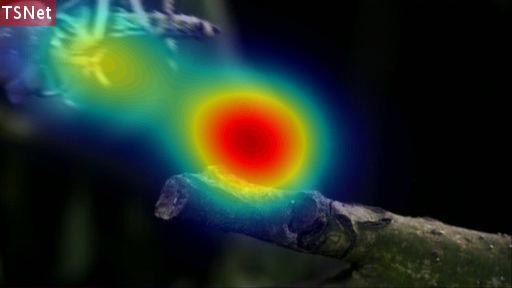} &			
		\includegraphics[width=0.135\textwidth]{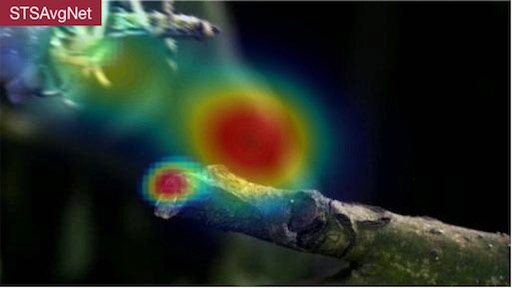} &	
		\includegraphics[width=0.135\textwidth]{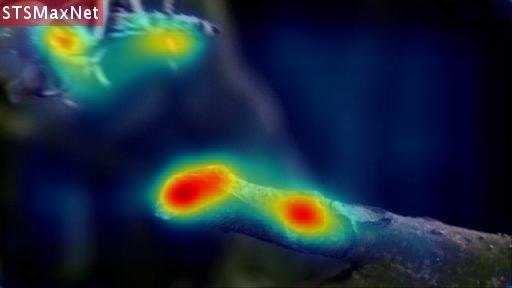} &	
		\includegraphics[width=0.135\textwidth]{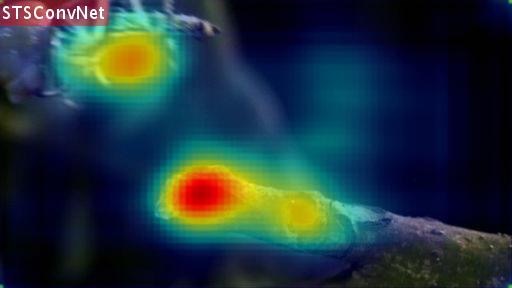} \\
		\includegraphics[width=0.135\textwidth]{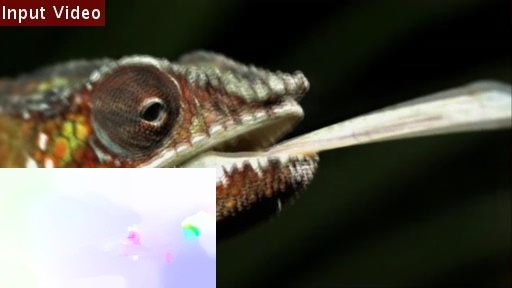} &
		\includegraphics[width=0.135\textwidth]{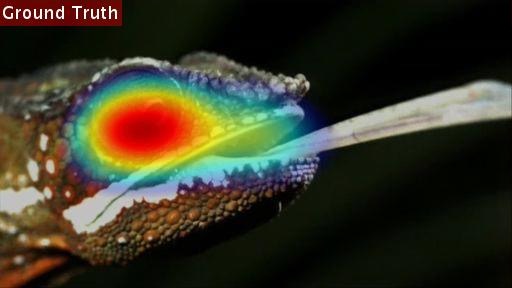} &	
		\includegraphics[width=0.135\textwidth]{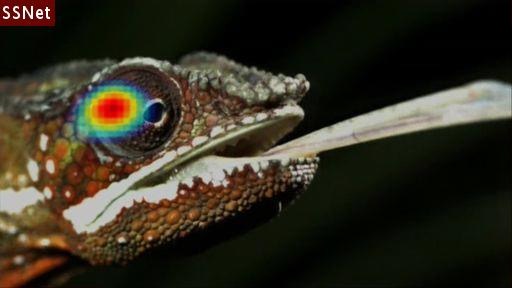} &	
		\includegraphics[width=0.135\textwidth]{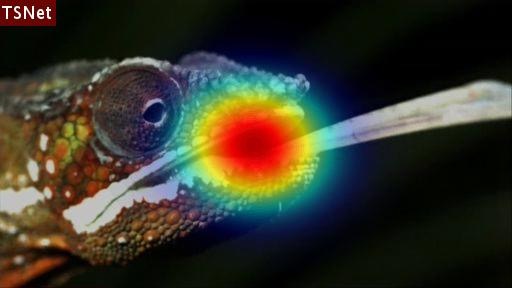} &	
		\includegraphics[width=0.135\textwidth]{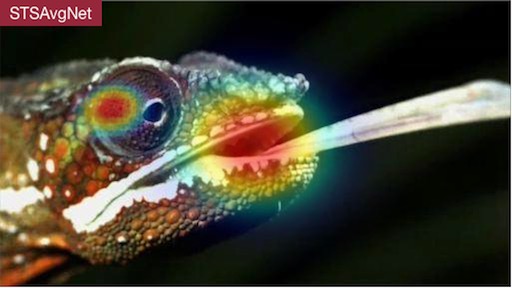} &	
		\includegraphics[width=0.135\textwidth]{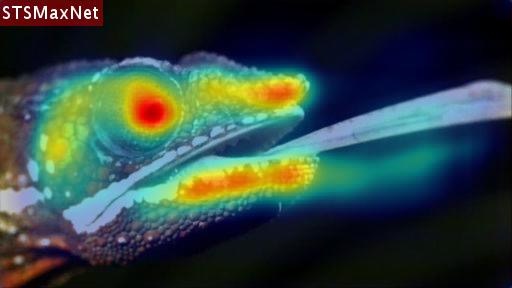} &	
		\includegraphics[width=0.135\textwidth]{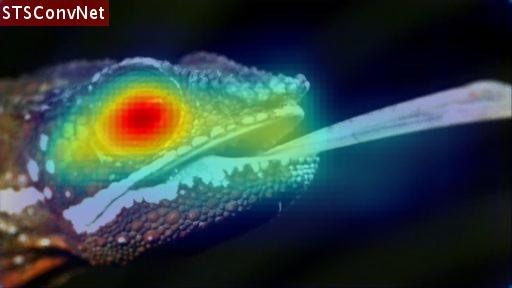} \\		
		\includegraphics[width=0.135\textwidth]{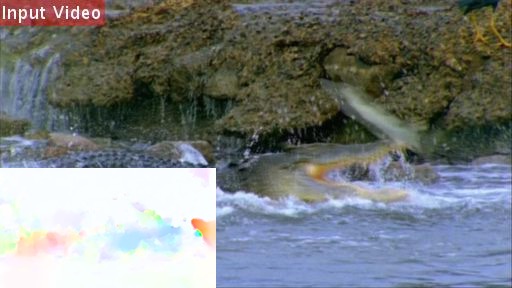} &
		\includegraphics[width=0.135\textwidth]{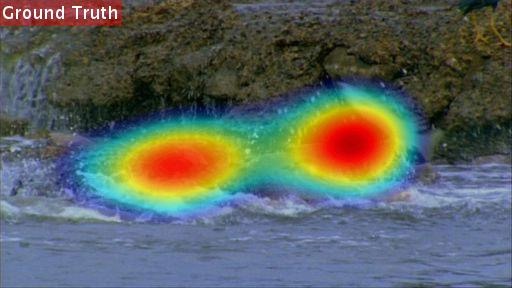} &				
		\includegraphics[width=0.135\textwidth]{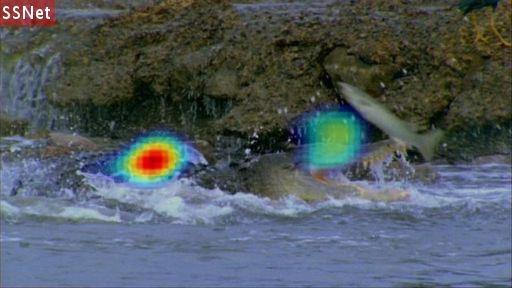} &	
		\includegraphics[width=0.135\textwidth]{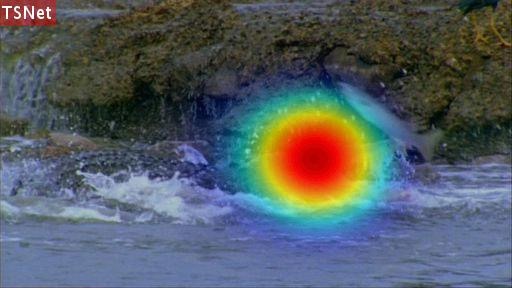} &		
		\includegraphics[width=0.135\textwidth]{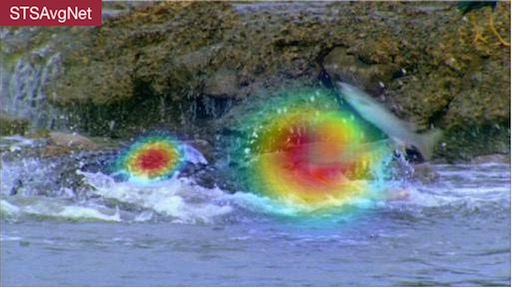} &		
		\includegraphics[width=0.135\textwidth]{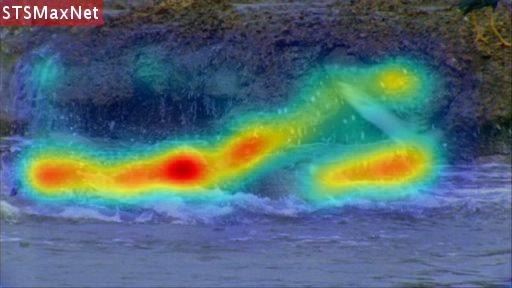} &	
		\includegraphics[width=0.135\textwidth]{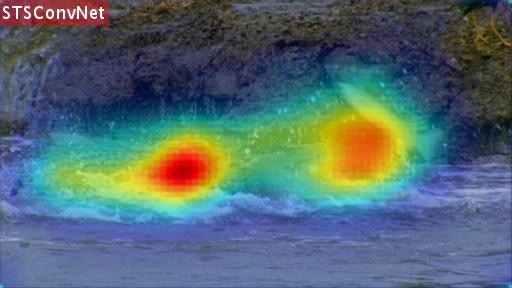} \\											
		\includegraphics[width=0.135\textwidth]{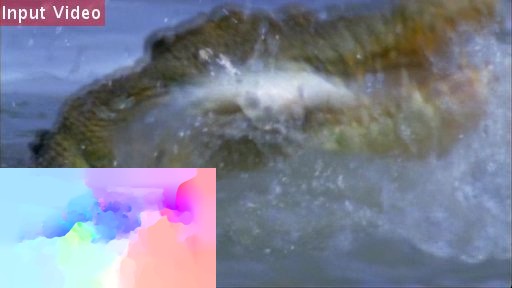} &
		\includegraphics[width=0.135\textwidth]{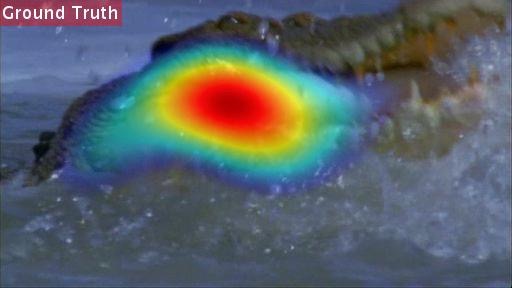} &				
		\includegraphics[width=0.135\textwidth]{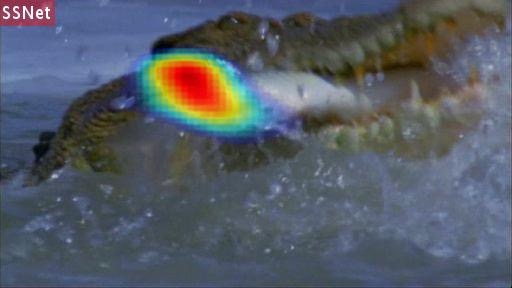} &	
		\includegraphics[width=0.135\textwidth]{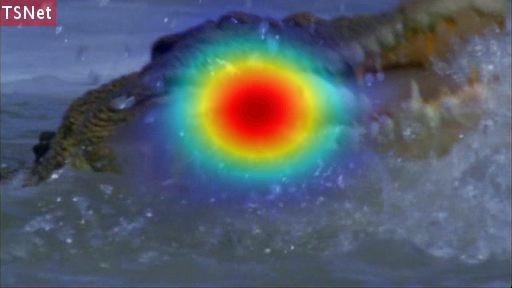} &		
		\includegraphics[width=0.135\textwidth]{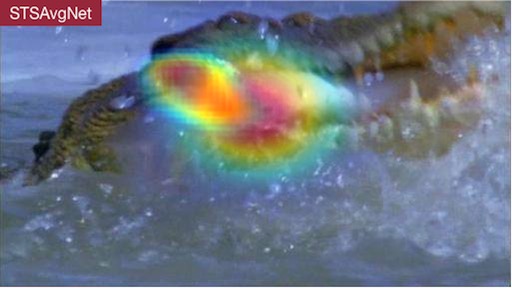} &		
		\includegraphics[width=0.135\textwidth]{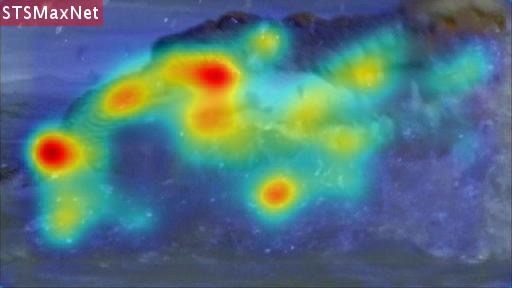} &		
		\includegraphics[width=0.135\textwidth]{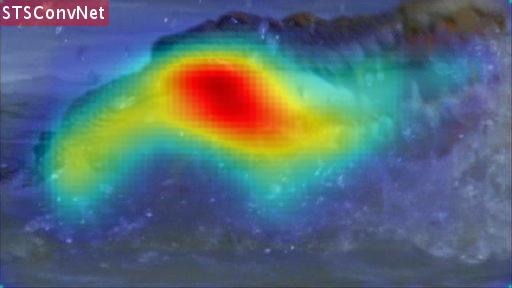} \\
		\small{Overlayed images} & \small{Ground Truth}  & \small SSNet & \small TSNet  & \small STSAvgNet & \small STSMaxNet & \small STSConvNet									
	\end{tabular}
	\caption{Qualitative evaluation of the proposed saliency network architectures. For these sample frames from the DIEM dataset, our STSConvNet provides the most accurate prediction as compared to the other network models.}
	\label{fig:network-comparisons}
\end{figure*}

\subsection{Experiments on DIEM}
In our analysis, we first both qualitatively and quantitatively compare our proposed deep dynamic saliency networks, SSNet, TSNet, STSDirectNet, STSAvgNet, STSMaxNet and STSConvNet, with each other. Following the experimental setup of~\cite{borji2013quantitative}, we split the dataset into a training set containing 64 video sequences and a testing set including the remaining 20 representative videos covering different concepts. Specifically, we use the same set of splits used in~\cite{rudoy2013learning}.  

As our STSMaxNet and STSConvNet models integrate spatial and temporal streams by respectively using element-wise and convolutional fusion strategies, we perform an extensive set of initial experiments to determine the optimum layers for the fusion process  to take place in these two-stream networks. In particular, we train STSMaxNet and STSConvNet models by considering different fusion layers, and test each one of them on the test set by considering sAUC measure.  In Table~\ref{tab:invest}, we provide these performance comparisons at different fusion layers. Interestingly, as can be seen from the table, fusing the spatial and temporal streams after the fifth convolution layer achieves the best results for both spatio-temporal networks. In the remainder, we use these settings for our STSMaxNet and STSConvNet models.

In Figure~\ref{fig:network-comparisons}, for some sample video frames we provide the outputs of the proposed networks along with the corresponding human fixation density maps (the ground truth). The input frames are given as overlayed images by compositing them with their consecutive frames to show the motion in the scenes. Saliency maps are shown as heatmaps superimposed over the original image for visualization purposes. As can be seen from these results, SSNet, which does employ appearance but not motion information, in general provides less accurate saliency maps and misses the foreground objects or their parts that are in motion. TSNet gives relatively better results, but as shown in the second and the third row, it does not identify all of the salient regions as it focuses more on the moving parts of the images. Directly averaging the saliency maps of these two single stream models, referred to as STSAvgNet, does not produce very satisfactory results either. As expected, STSMaxNet is slightly better since max fusion enforces to learn more effective filters in order to combine the spatial and temporal streams. Overall, STSConvNet is the best performing model. This can be rooted in the application of $1 \times 1$ convolutional filters that learn the optimal weights to combine appearance and motion feature maps.

\begingroup
\begin{table}[!t]
	\centering
		\caption{Performance comparison of our spatio-temporal saliency networks at different fusion layers on the DIEM dataset. The reported values are sAUC scores. Best performance is achieved after the fifth convolution layer.}
	
	\begin{tabular}{|ccc|}
		\hline
		Fusion Layers & STSMaxNet & STSConvNet\\
		\hline \hline
		Conv2 & 0.52 & 0.71\\
		Conv3 & 0.67 & 0.70\\
		Conv4 & 0.76 & 0.83\\
		Conv5 & 0.81 & 0.84\\
		Conv6 & 0.80 & 0.79\\
		Conv7 & 0.81 & 0.79\\
		\hline
	\end{tabular}
	\label{tab:invest}
\end{table}
\endgroup

\renewcommand*{\arraystretch}{1.2}
\begin{table}[!t]
	\centering
		\caption{Performance comparisons on the DIEM dataset.}
	\begin{tabular}{|p{2.25cm}cccccc|}
		\hline
		& AUC & sAUC & CC & NSS & ${\chi}^2$ & NCC\\
		\hline \hline
		SSNet & 0.72 & 0.69 & 0.35 & 1.85 & 0.48 & 0.41\\
		TSNet  & 0.79 & 0.77 & 0.41 & 1.98 & 0.40 & 0.43\\
		\hline
			STSDirectNet & 0.71 & 0.60 & 0.37 & 1.53 & 0.49 & 0.28\\
		STSAvgNet & 0.68 & 0.62 & 0.37 & 1.67 & 0.49 & 0.37\\
		STSMaxNet & 0.83 & 0.81 & 0.46 & 2.01 & 0.31 & 0.45\\
		STSConvNet & 0.87 & 0.84 & 0.47 & 2.15 & 0.29 & 0.46\\
		STSConvNet* & \textbf{0.88} & \textbf{0.86} & \textbf{0.48} & 2.18 & \textbf{0.28} & \textbf{0.47}\\
		\hline 
		GBVS~\cite{harel2006graph} & 0.74 & 0.70 & 0.47 & 2.04 & 0.47 & 0.38\\
		SR~\cite{hou2007saliency} & 0.69 & 0.64 & 0.30 & 2.22 & 0.57 & 0.40\\
		PQFT~\cite{guo2008spatio} & 0.71 & 0.67 & 0.33 & 1.77 & 0.52 & 0.33\\
		Seo-Milanfar~\cite{seo2009static} & 0.59 & 0.51 & 0.10 & 0.12 & 0.75 & 0.28\\
		Rudoy \emph{et al.}~\cite{rudoy2013learning} & -- & 0.74 & -- & -- & 0.31 & --\\
		Fang \emph{et al.}~\cite{fang2014} & 0.71 & 0.48 & 0.21 & 0.55 & 0.87 & 0.40\\
		Zhou \emph{et al.}~\cite{zhou2014learning} & 0.60 & 0.52 & 0.13 & 0.24 & 0.71 & 0.22\\
		DWS~\cite{awsd} & 0.81 & 0.79 & 0.32 & \textbf{2.97} & 0.40 & 0.39\\
		\hline
	\end{tabular}
	\label{tab:DIEMresults}
\end{table}

\begin{figure*}[!t]
	\centering
	\begin{tabular}{c@{\;}c@{\;}c@{\;}c@{\;}c@{\;}c@{\;}c}
		\includegraphics[width=0.135\textwidth]{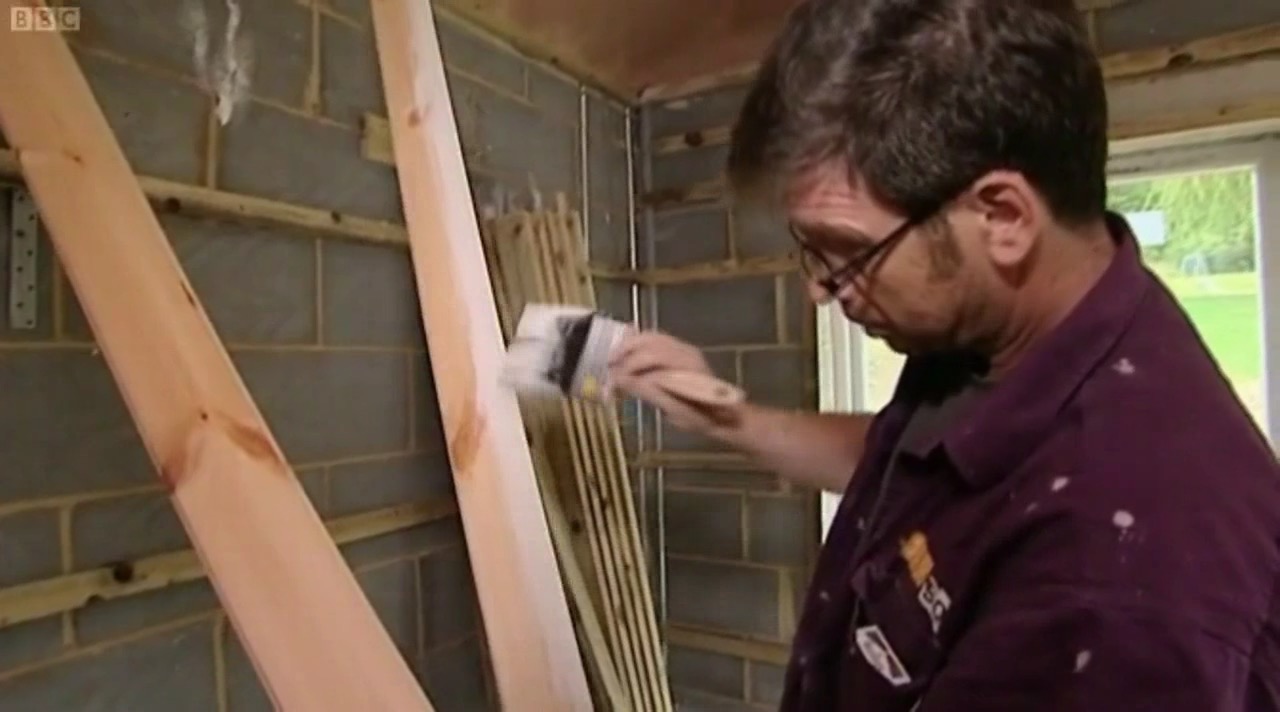} &
		\includegraphics[width=0.135\textwidth]{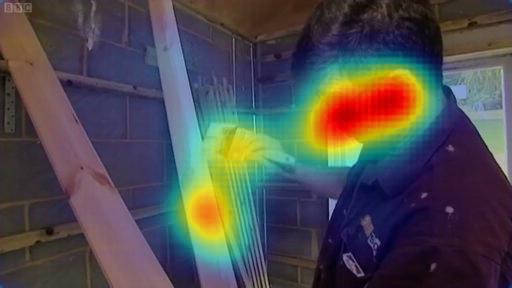} &
		\includegraphics[width=0.135\textwidth]{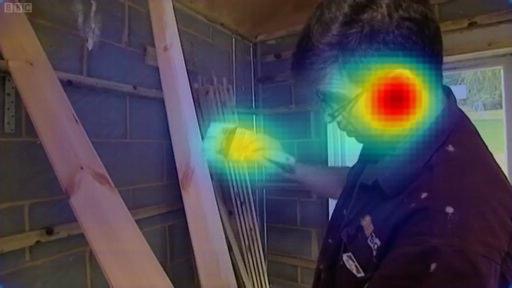} &
		\includegraphics[width=0.135\textwidth]{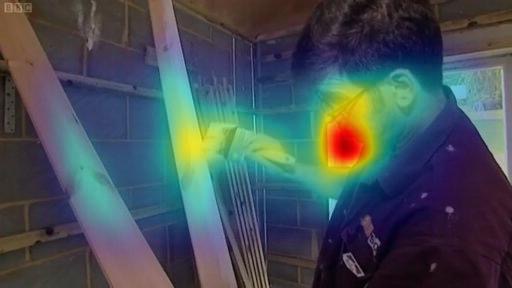} &
		\includegraphics[width=0.135\textwidth]{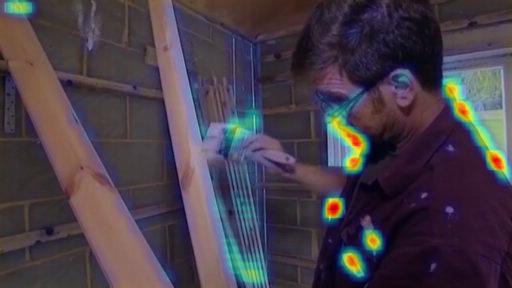} &
		\includegraphics[width=0.135\textwidth]{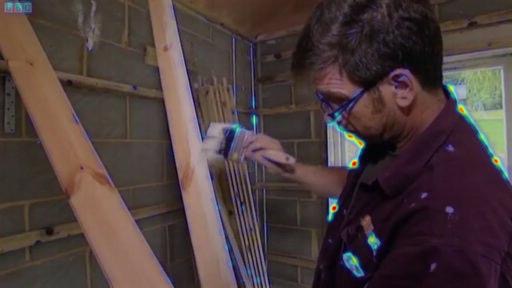} &
		\includegraphics[width=0.135\textwidth]{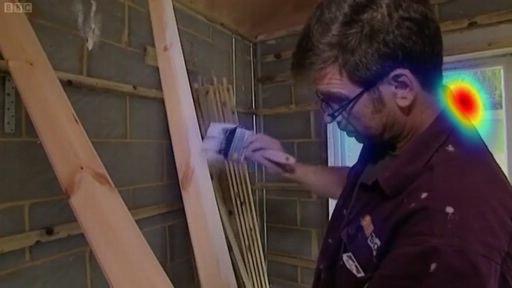} \\		
		
		\includegraphics[width=0.135\textwidth]{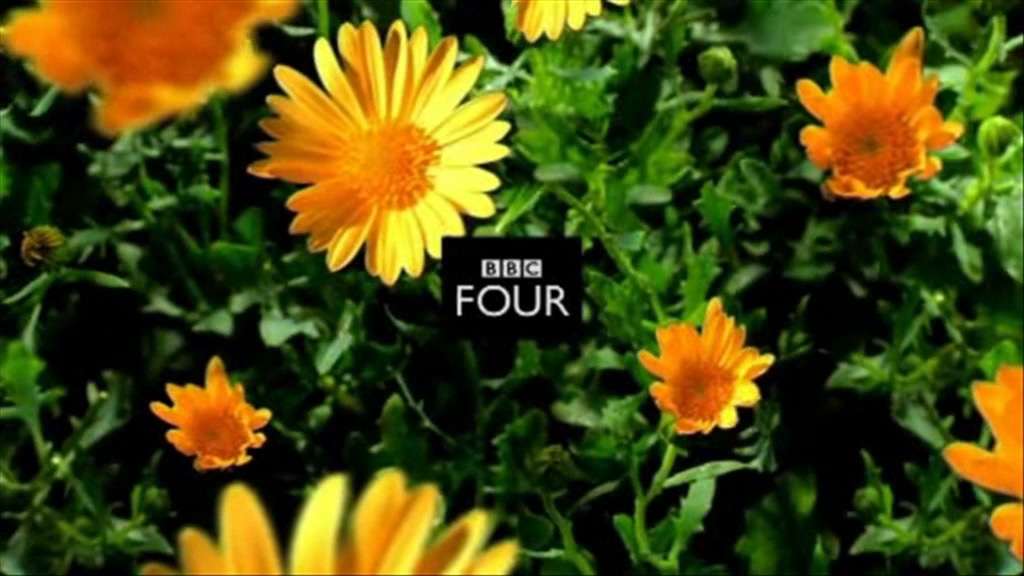} &
		\includegraphics[width=0.135\textwidth]{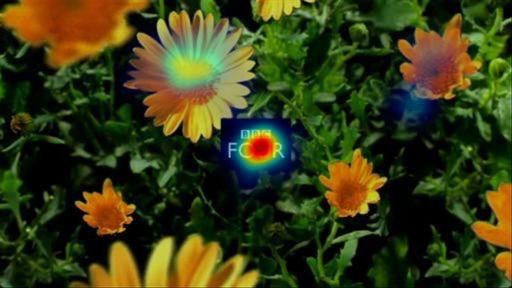} &
		\includegraphics[width=0.135\textwidth]{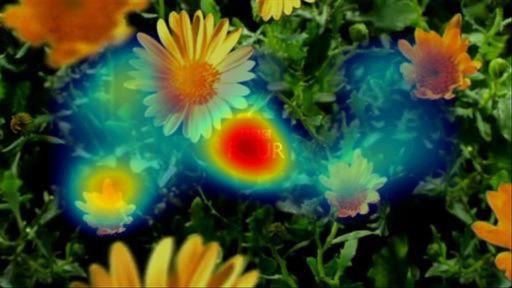} &
		\includegraphics[width=0.135\textwidth]{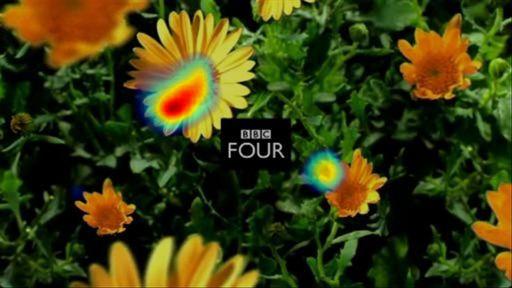} &
		\includegraphics[width=0.135\textwidth]{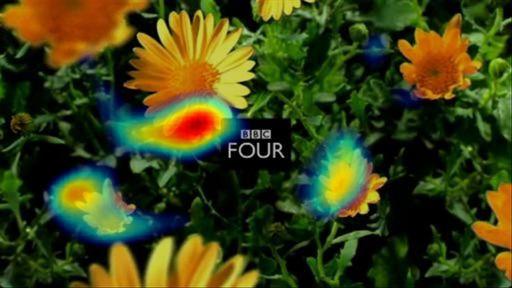} &
		\includegraphics[width=0.135\textwidth]{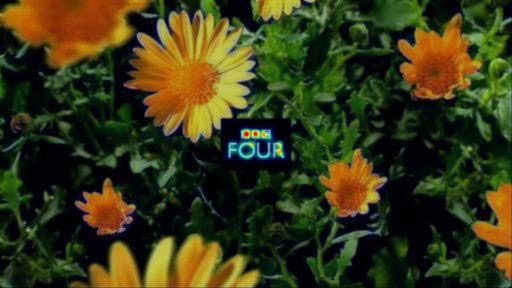} &
		\includegraphics[width=0.135\textwidth]{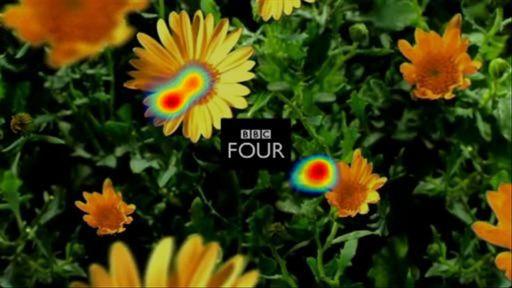} \\
		
		\includegraphics[width=0.135\textwidth]{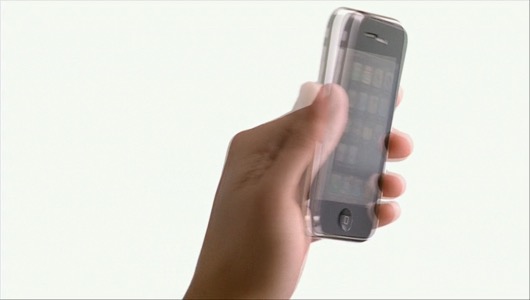} &
		\includegraphics[width=0.135\textwidth]{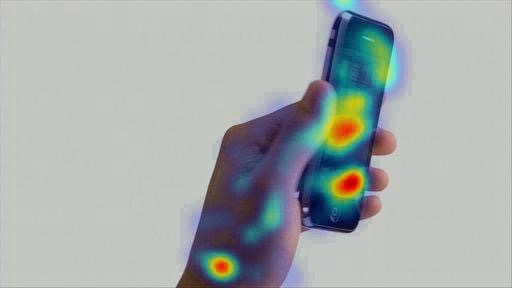} &
		\includegraphics[width=0.135\textwidth]{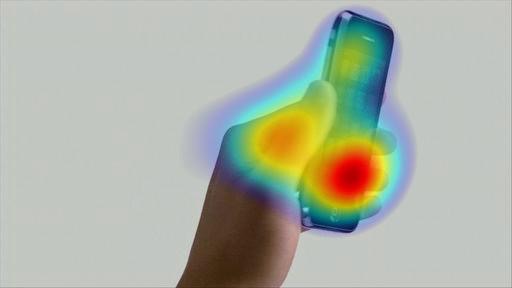} &
		\includegraphics[width=0.135\textwidth]{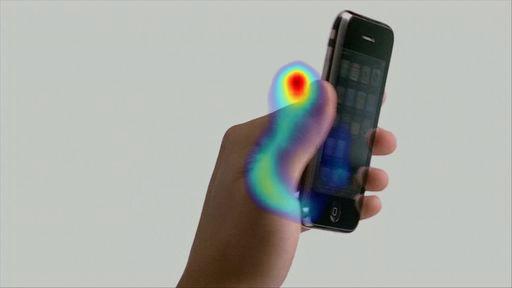} &
		\includegraphics[width=0.135\textwidth]{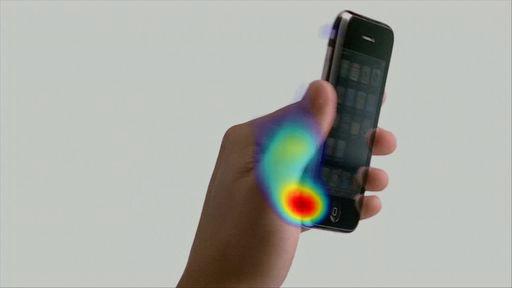} &
		\includegraphics[width=0.135\textwidth]{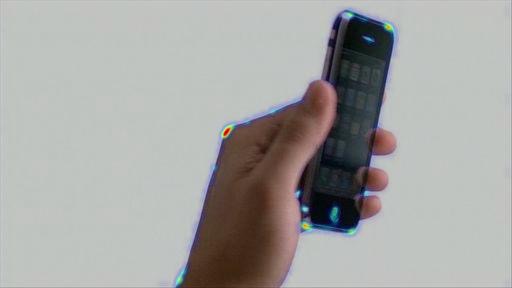}&
		\includegraphics[width=0.135\textwidth]{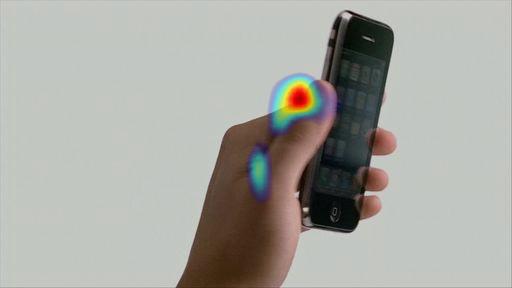} \\
		
		\includegraphics[width=0.135\textwidth]{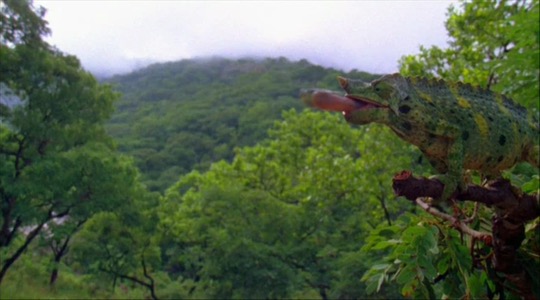} &
		\includegraphics[width=0.135\textwidth]{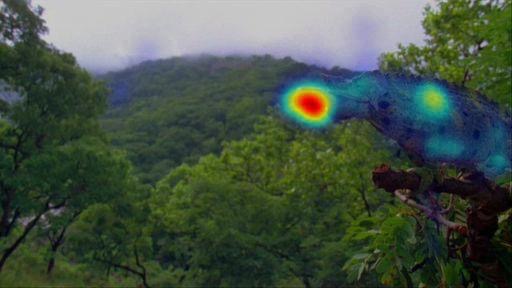} &
		\includegraphics[width=0.135\textwidth]{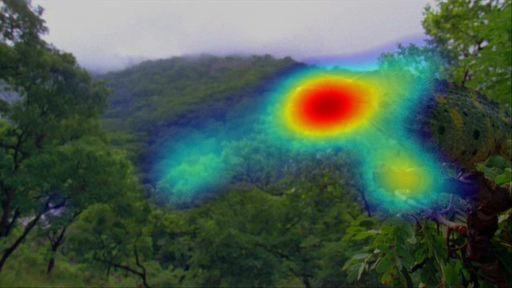} &
		\includegraphics[width=0.135\textwidth]{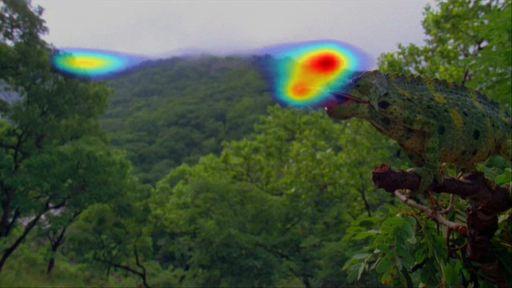} &
		\includegraphics[width=0.135\textwidth]{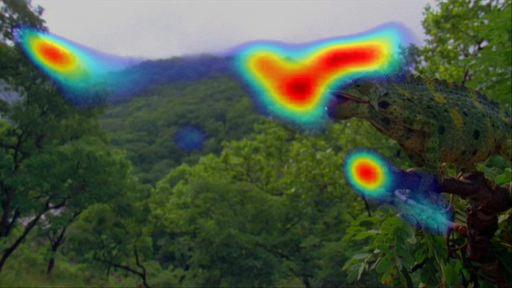} &
		\includegraphics[width=0.135\textwidth]{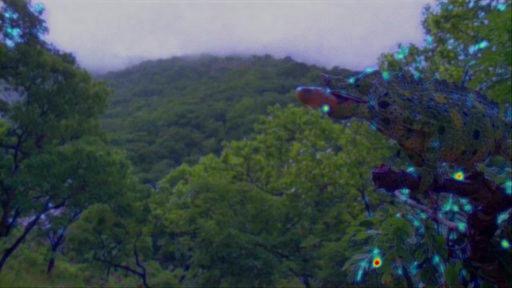} &
		\includegraphics[width=0.135\textwidth]{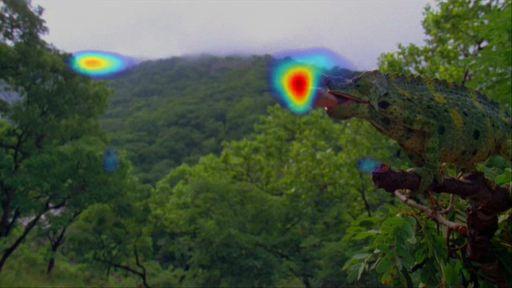} \\
		
		\includegraphics[width=0.135\textwidth]{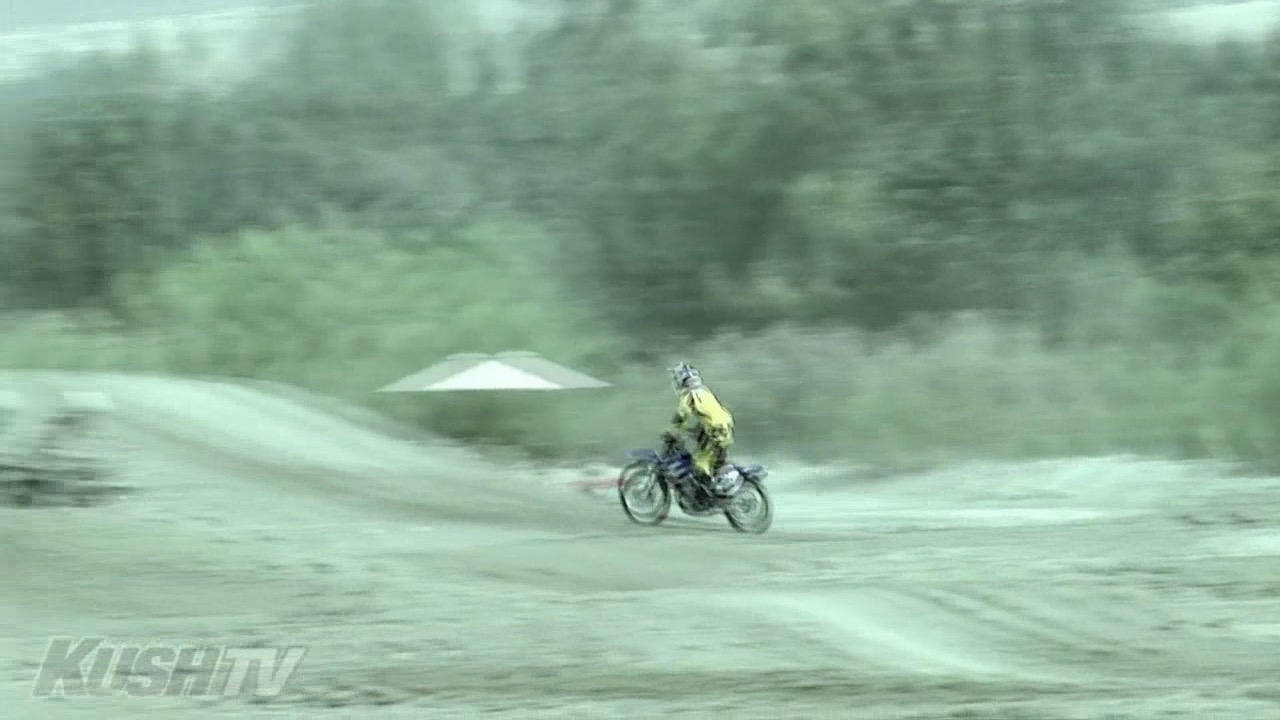} &
		\includegraphics[width=0.135\textwidth]{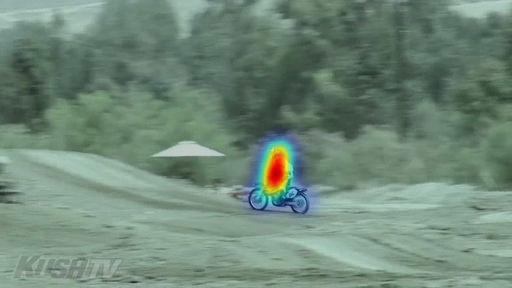} &
		\includegraphics[width=0.135\textwidth]{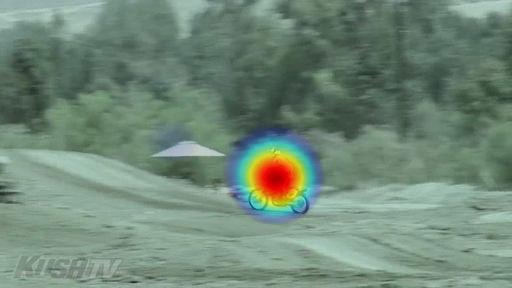} &
		\includegraphics[width=0.135\textwidth]{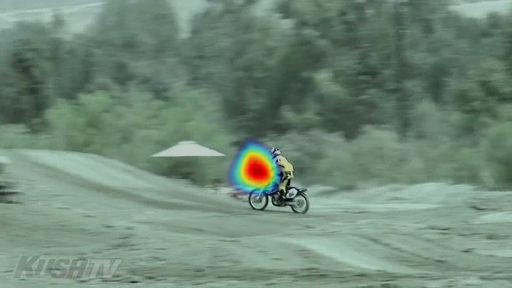} &
		\includegraphics[width=0.135\textwidth]{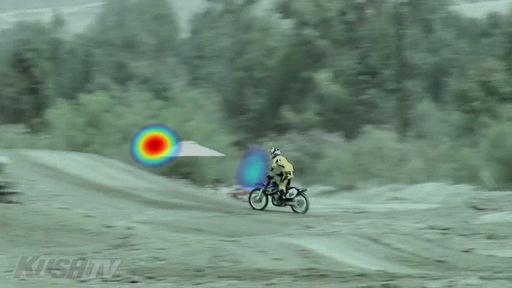} &
		\includegraphics[width=0.135\textwidth]{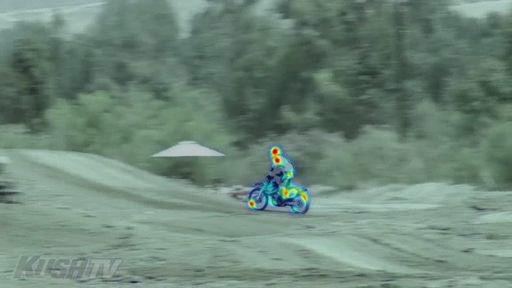} &
		\includegraphics[width=0.135\textwidth]{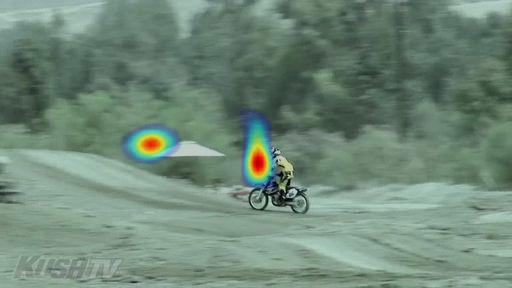} \\
		
		\includegraphics[width=0.135\textwidth]{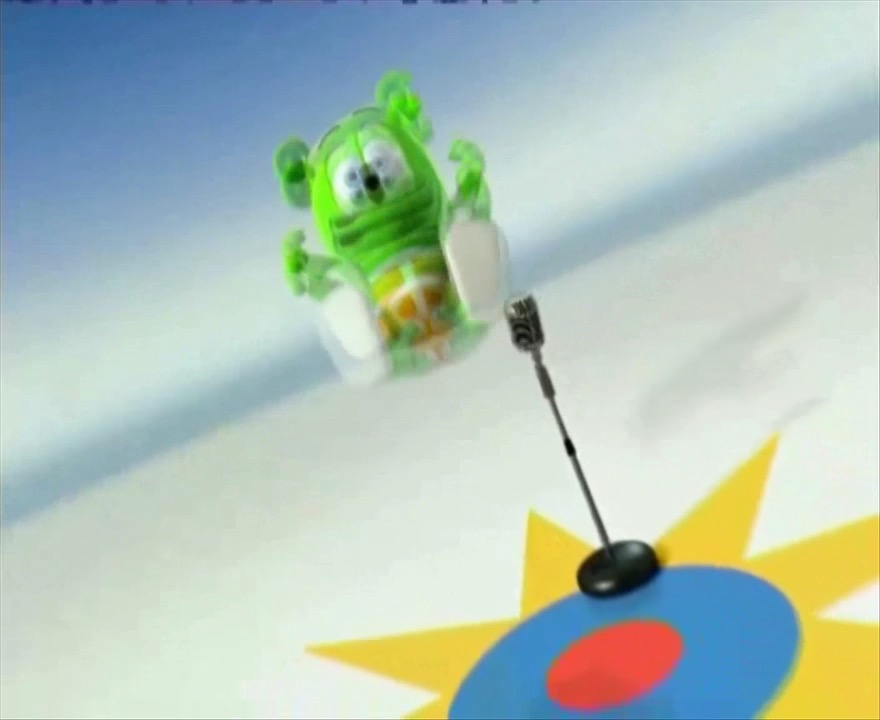} &
		\includegraphics[width=0.135\textwidth]{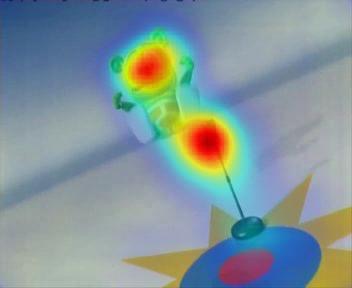} &
		\includegraphics[width=0.135\textwidth]{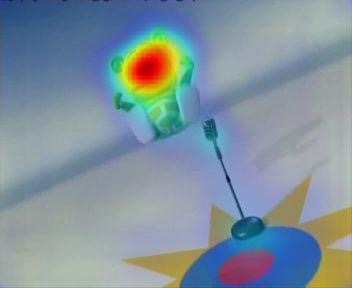} &
		\includegraphics[width=0.135\textwidth]{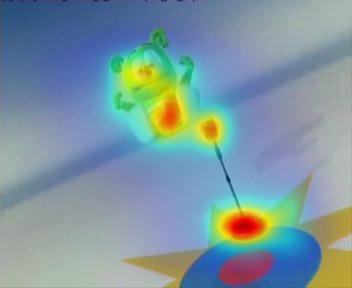} &
		\includegraphics[width=0.135\textwidth]{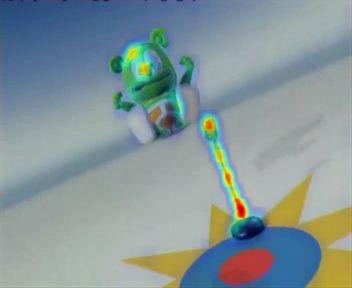} &
		\includegraphics[width=0.135\textwidth]{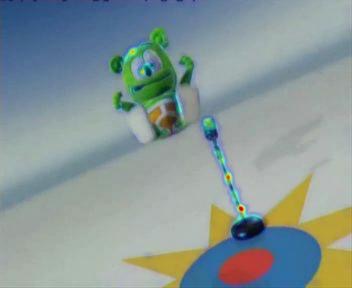} &
		\includegraphics[width=0.135\textwidth]{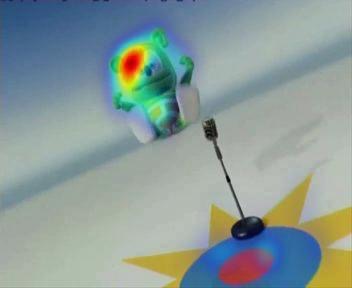} \\
		
		\small Overlayed frames &
		\small Ground Truth & \small STSConvNet* & \small GBVS~\cite{harel2006graph} & \small PQFT~\cite{guo2008spatio} & \small SR~\cite{hou2007saliency} & \small DWS~\cite{awsd}
	\end{tabular}
	\caption{Qualitative comparison of our STSConvNet* model against some dynamic saliency models on DIEM dataset. Our model clearly produces better results.}
	\label{fig:tab:model-comparisons}
\end{figure*}

In Table~\ref{tab:DIEMresults}, we present the quantitative results averaged over all video sequences and frames on the test split of the DIEM dataset. Here, we compare and contrast our single- and two-stream saliency networks with eight existing dynamic saliency methods, namely GVBS~\cite{harel2006graph}, SR~\cite{hou2007saliency}, PQFT~\cite{guo2008spatio}, Seo and Milanfar~\cite{seo2009static}, Rudoy \emph{et al.}~\cite{rudoy2013learning}\footnote{Since the code provided by the authors is not working correctly, sAUC and $\chi^2$ scores are directly taken from~\cite{rudoy2013learning}.}, Fang \emph{et al.}~\cite{fang2014}, Zhou \emph{et al.}~\cite{zhou2014learning} and DWS~\cite{awsd} models.

Among our deep saliency networks, we empirically find that STSDirectNet provides the worst quantitative results. This is in line with our observation in Table~\ref{tab:invest} that delaying the integration of appearance and motion streams to a certain extent leads to more effective learning of mid and low level features. Secondly, we see that SSNet performs considerably lower than Temporal stream network, which demonstrates that motion is more vital for dynamic saliency. STSMaxNet gives results better than those of the single stream models but our STSConvNet model performs even better. It can be argued that STSConvNet learns more effective filters that combine spatial and temporal streams in an optimal manner. In addition, when we employ the data augmentation strategy proposed in the previous section, it further improves the overall performance of STSConvNet. In the remainder, we refer to this model with data augmentation as STSConvNet*. When we compare our proposed STSMaxNet, STSConvNet, and  STSConvNet* models to the previous dynamic saliency methods, our results demonstrate the advantages of two-stream deep CNNs that they consistently outperform all those approaches, including the very recently proposed DWS model, according to five out of six evaluation measures. 

We present some qualitative results in Figure~\ref{fig:tab:model-comparisons} where we again provide the input frames as transparent overlayed images showing the inherent motion. We observe that the proposed STSConvNet* model localizes the salient regions more accurately than the existing models. For example, for the frame given in the first row, none of the compared models correctly capture the fixations over the painting brush. Similarly, for the second and the third frames, only our spatio-temporal saliency network fixates to the text and the cellular phone in the frames, respectively. 

\subsection{Experiments on UCF-Sports}

Learning-based models might sometimes fail to provide satisfactory results for a test sample due to a shift from the training data domain. To validate generalization ability of our best-performing STSConvNet* model, we perform additional experiments on UCF-Sports dataset. In particular, we do not carry out any training for our model from scratch or fine-tune it on UCF-Sports but rather use the predictions of the model trained only on DIEM dataset.

\begingroup
\renewcommand*{\arraystretch}{1.2}
\begin{table}[!t]
	\centering
		\caption{Performance comparisons on the UCF-SPORTS dataset.}
	\begin{tabular}{|p{2.25cm}cccccc|}
		\hline
		& AUC & sAUC & CC & NSS & ${\chi}^2$ & NCC\\
		\hline \hline
		GBVS~\cite{harel2006graph} &  0.83 & 0.52 & 0.46 & 1.82 & 0.54 & \textbf{0.59}\\
		SR~\cite{hou2007saliency} &  0.78 & 0.69 & 0.26 & 1.20 & 0.42 & 0.52\\
		PQFT~\cite{guo2008spatio} & 0.69 & 0.51 & 0.29 & 1.15 & 0.64 & 0.48\\
		Seo-Milanfar~\cite{seo2009static} & 0.80 & 0.72 & 0.31 & 1.37 & 0.56 & 0.36\\
		Fang \emph{et al.}~\cite{fang2014} & \textbf{0.85} & 0.70 & 0.44 & 1.95 & 0.52 & 0.33\\
		Zhou \emph{et al.}~\cite{zhou2014learning} & 0.81 & 0.72 & 0.36 & 1.71 & 0.56 & 0.37\\
		DWS~\cite{awsd} & 0.76 & 0.70 & 0.28 & 2.01 & 0.40 & 0.49\\
		STSConvNet* & 0.82 & \textbf{0.75} & \textbf{0.48} & \textbf{2.13} & \textbf{0.39} & 0.54\\
		\hline
	\end{tabular}
	\label{tab:ucf-results}
\end{table}
\endgroup

\begin{figure*}[!t]
	\centering
	\resizebox{\textwidth}{!} {
	\begin{tabular}{c@{\;}c@{\;}c@{\;}c@{\;}c@{\;}c}
		\includegraphics[width=0.18\textwidth]{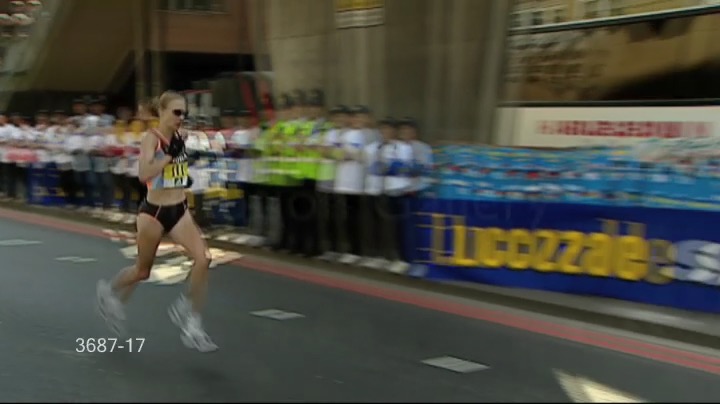} &
		\includegraphics[width=0.18\textwidth]{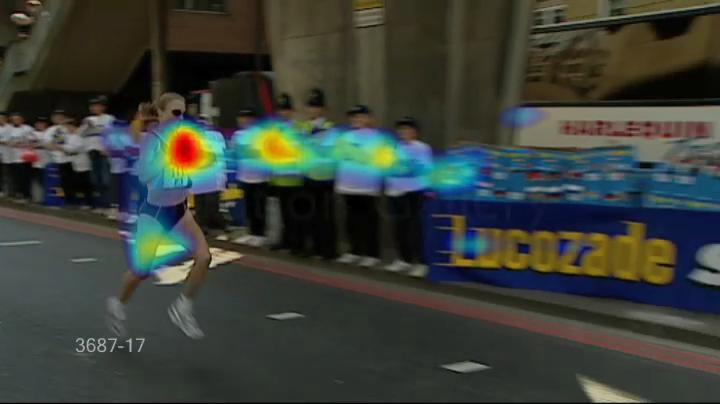} &
		\includegraphics[width=0.18\textwidth]{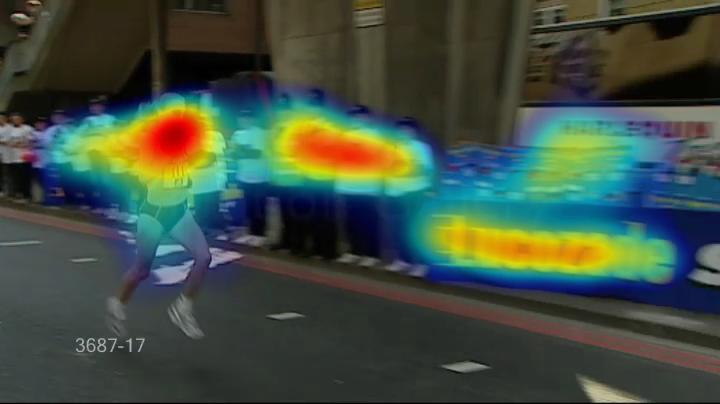} &
		\includegraphics[width=0.18\textwidth]{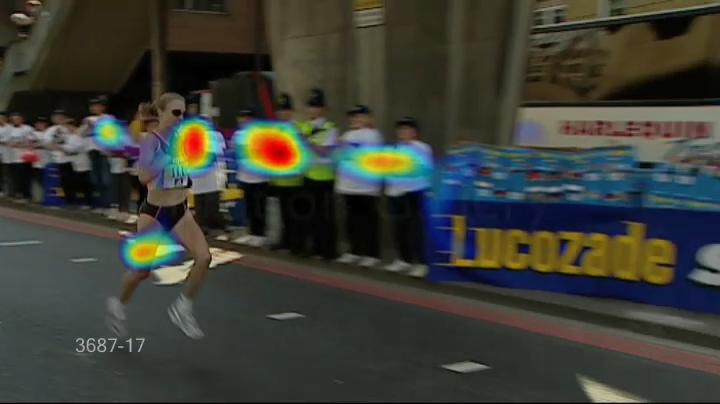} &
		\includegraphics[width=0.18\textwidth]{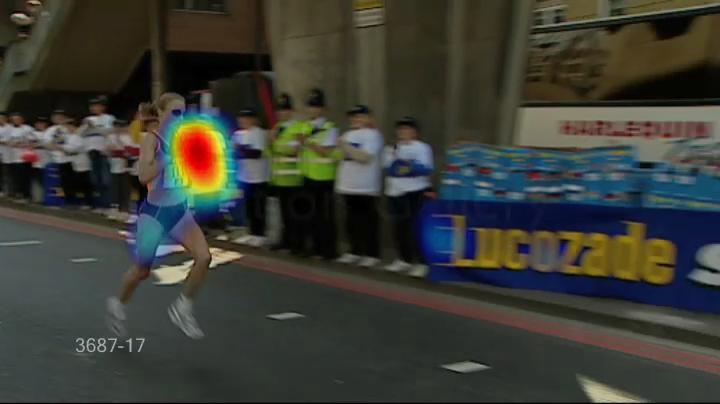} \\	
		\includegraphics[width=0.18\textwidth]{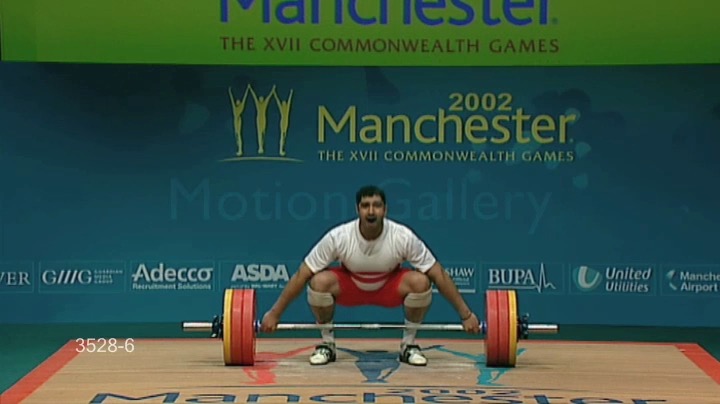} &
		\includegraphics[width=0.18\textwidth]{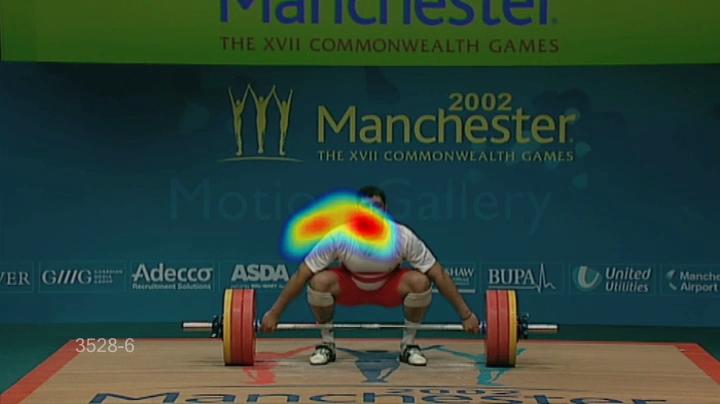} &
		\includegraphics[width=0.18\textwidth]{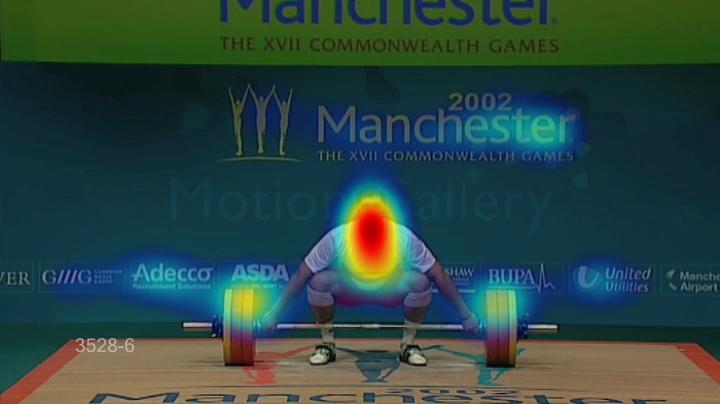} &
		\includegraphics[width=0.18\textwidth]{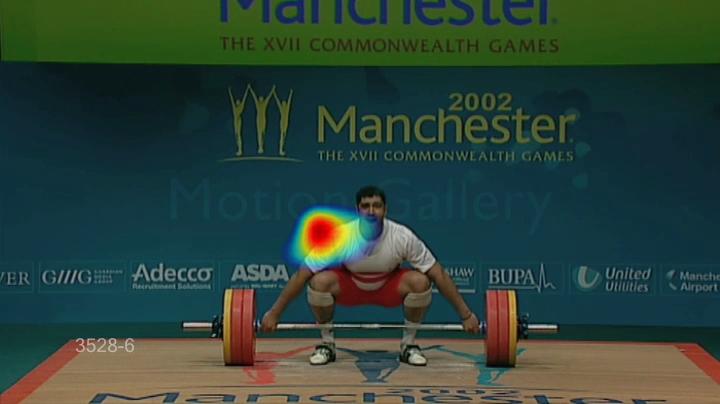} &
		\includegraphics[width=0.18\textwidth]{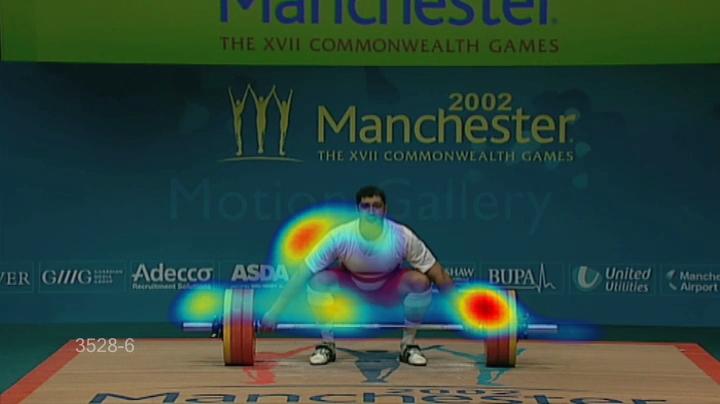} \\
		\includegraphics[width=0.18\textwidth]{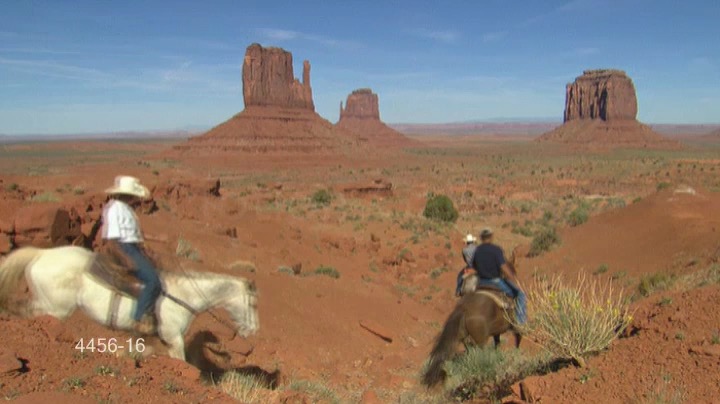} &
		\includegraphics[width=0.18\textwidth]{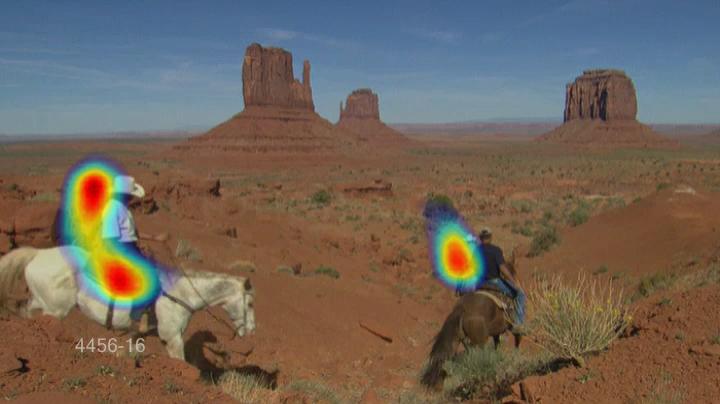} &
		\includegraphics[width=0.18\textwidth]{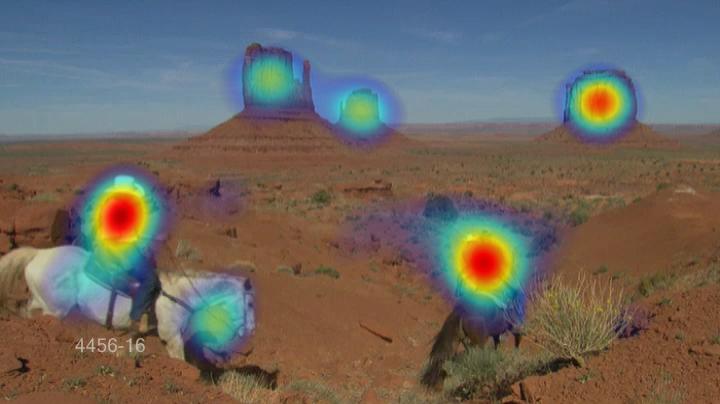} &
		\includegraphics[width=0.18\textwidth]{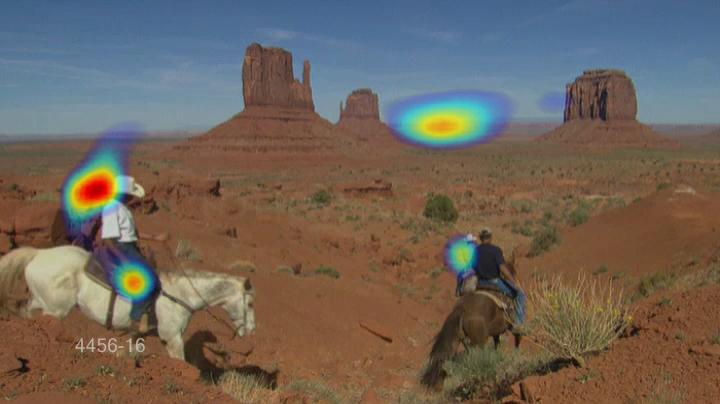} &
		\includegraphics[width=0.18\textwidth]{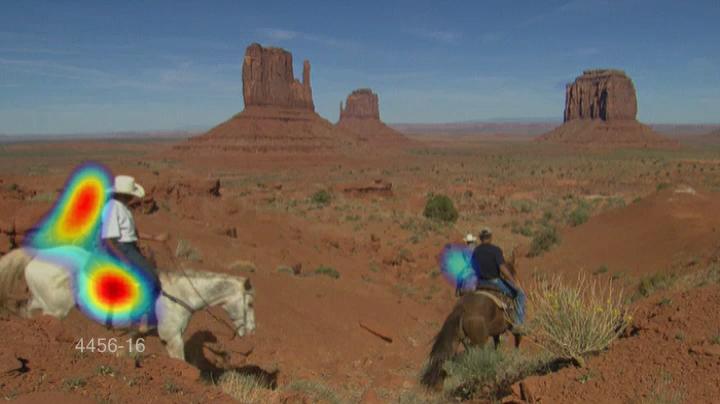} \\		

		\includegraphics[width=0.18\textwidth]{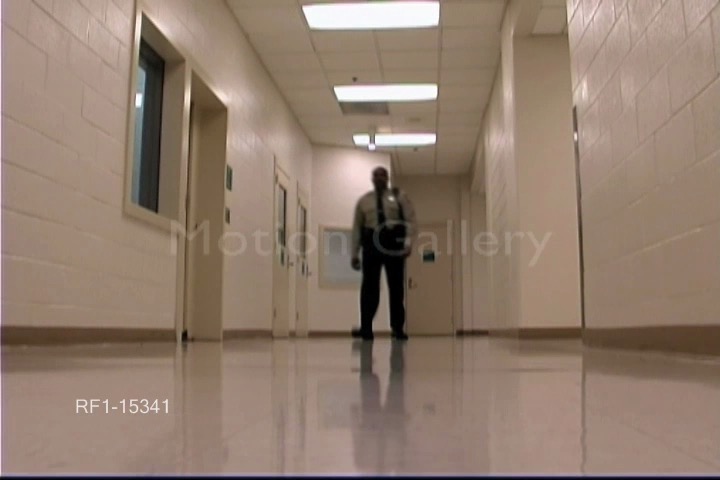} &
		\includegraphics[width=0.18\textwidth]{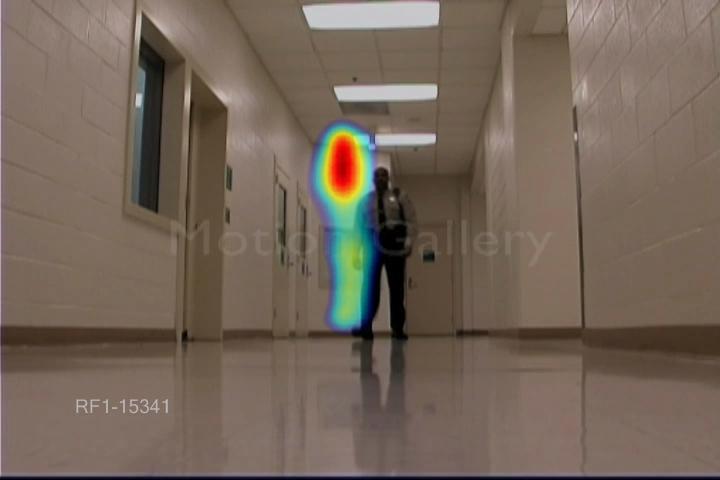} &
		\includegraphics[width=0.18\textwidth]{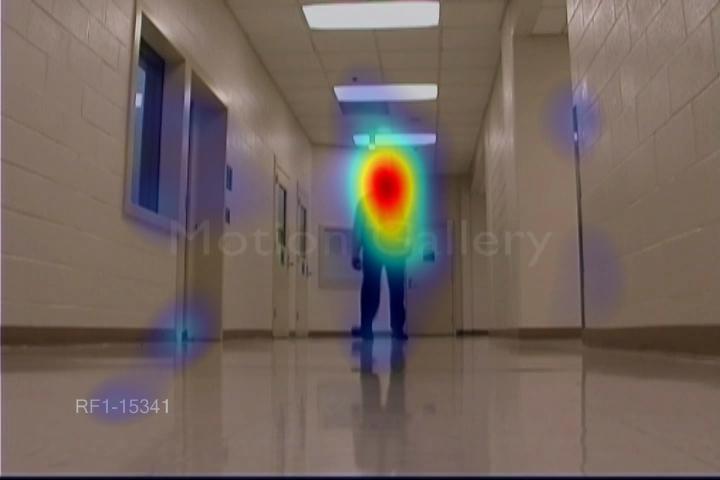} &
		\includegraphics[width=0.18\textwidth]{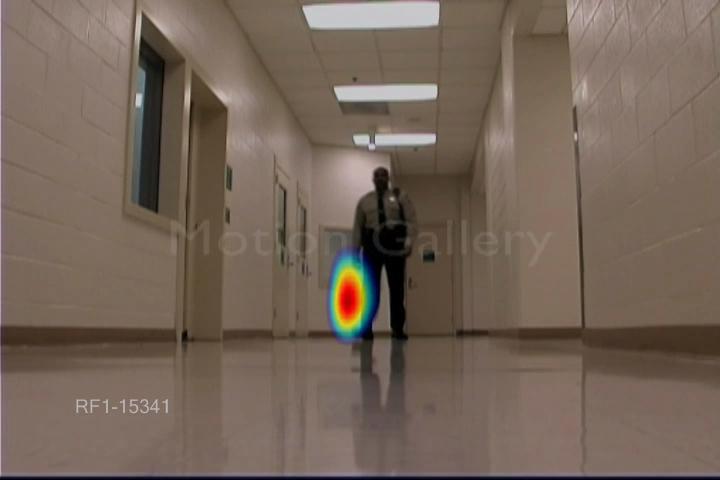} &
		\includegraphics[width=0.18\textwidth]{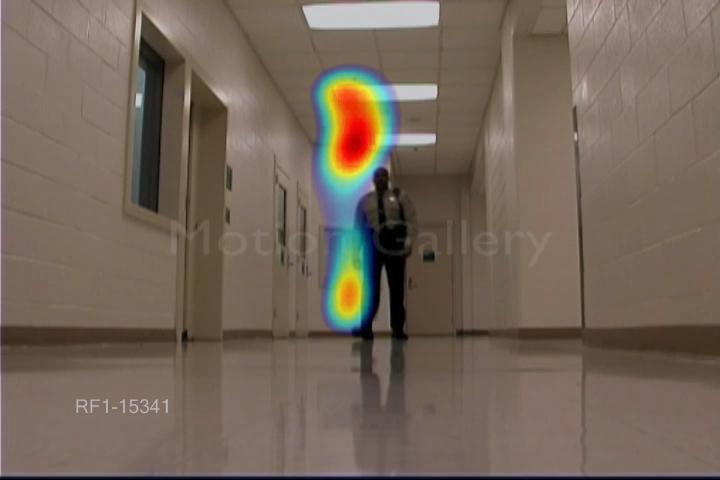} \\
		\includegraphics[width=0.18\textwidth]{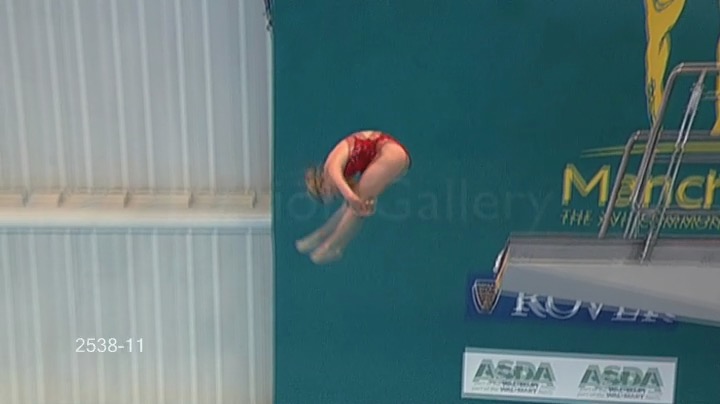} &	
		\includegraphics[width=0.18\textwidth]{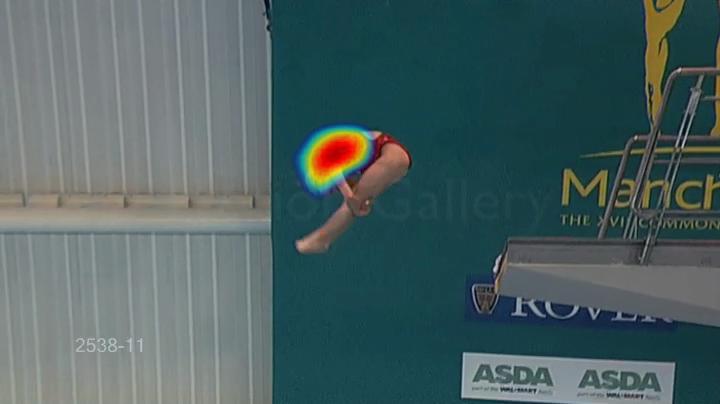} &
		\includegraphics[width=0.18\textwidth]{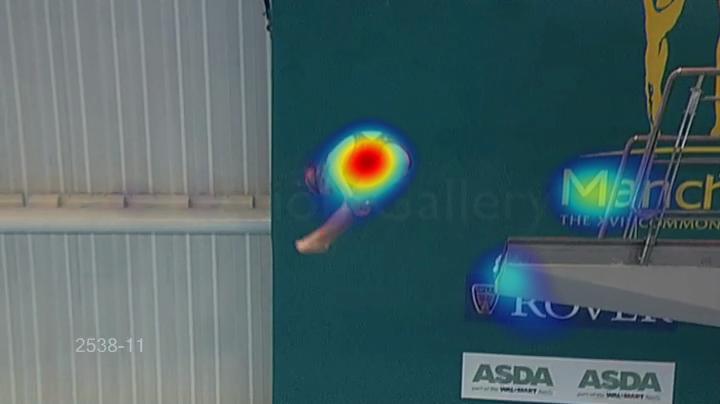} &
		\includegraphics[width=0.18\textwidth]{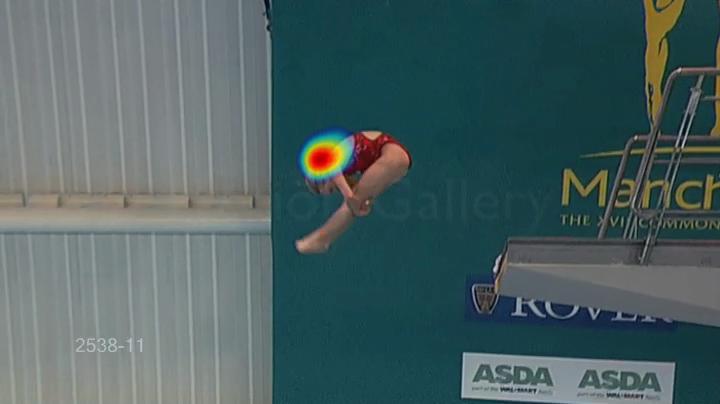} &
		\includegraphics[width=0.18\textwidth]{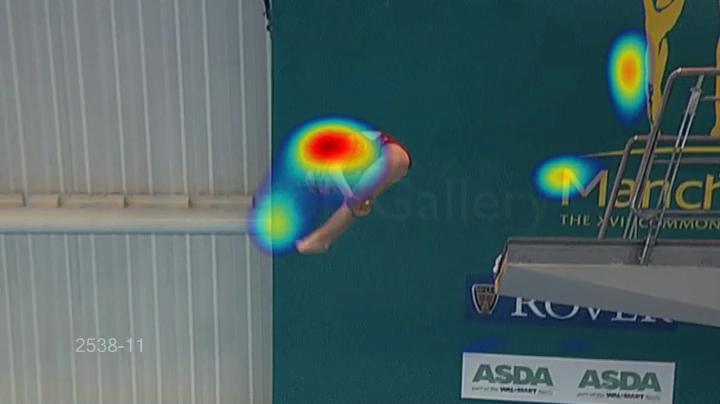} \\
		
		\footnotesize Overlayed input frames & \footnotesize Ground Truth & \footnotesize STSConvNet* & \footnotesize Fang \emph{et al.}~\cite{fang2014} & \footnotesize DWS~\cite{awsd}
	\end{tabular}}
	\caption{Qualitative comparison of our STSConvNet* model against some previous dynamic saliency models on UCF-Sports dataset. Our spatio-temporal saliency network outperforms the others.}
	\label{fig:tab:early-ucf}
\end{figure*}

\begin{figure}[!t]
	\centering
	\resizebox{\linewidth}{!} {
	\begin{tabular}{c@{\;}c@{\;}c@{\;}c@{\;}c}
		\includegraphics[width=2.4cm,height=1.8cm]{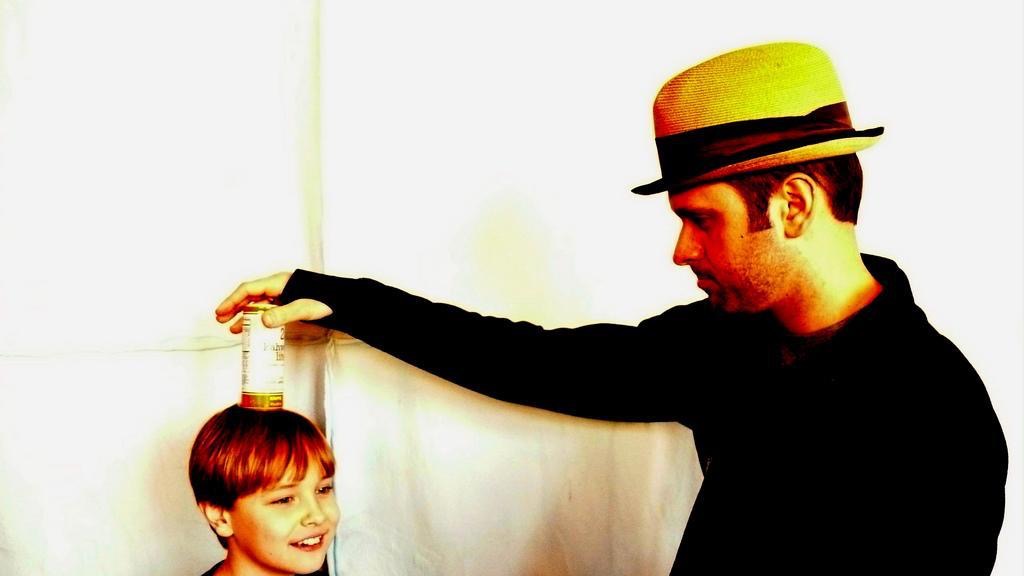} &
		\includegraphics[width=2.4cm,height=1.8cm]{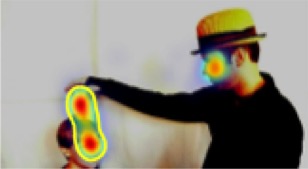} &
		\includegraphics[width=2.4cm,height=1.8cm]{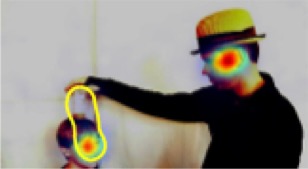} &
		\includegraphics[width=2.4cm,height=1.8cm]{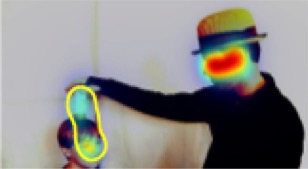} &
		\includegraphics[width=2.4cm,height=1.8cm]{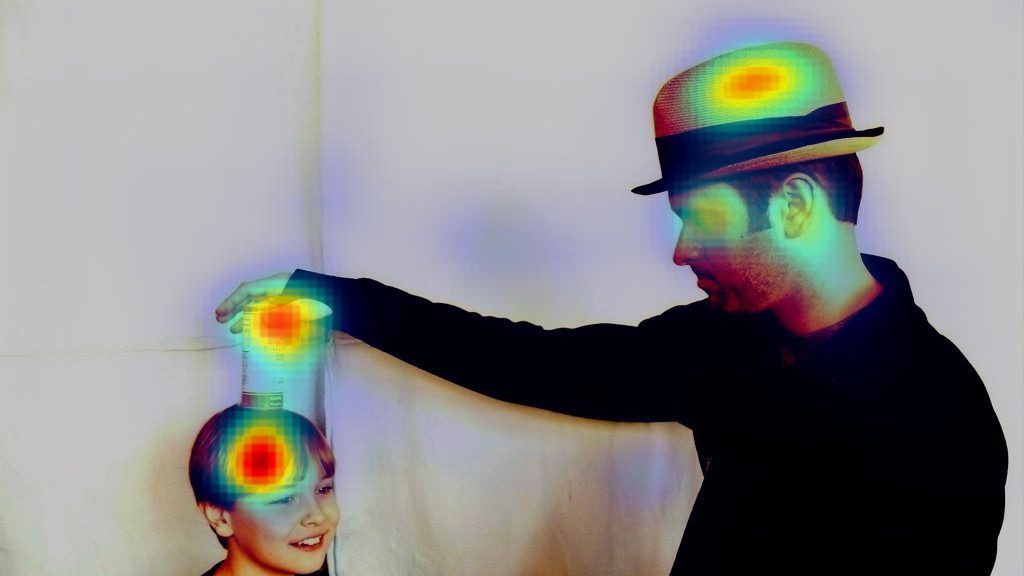} \\
		\includegraphics[width=2.4cm,height=3.4cm]{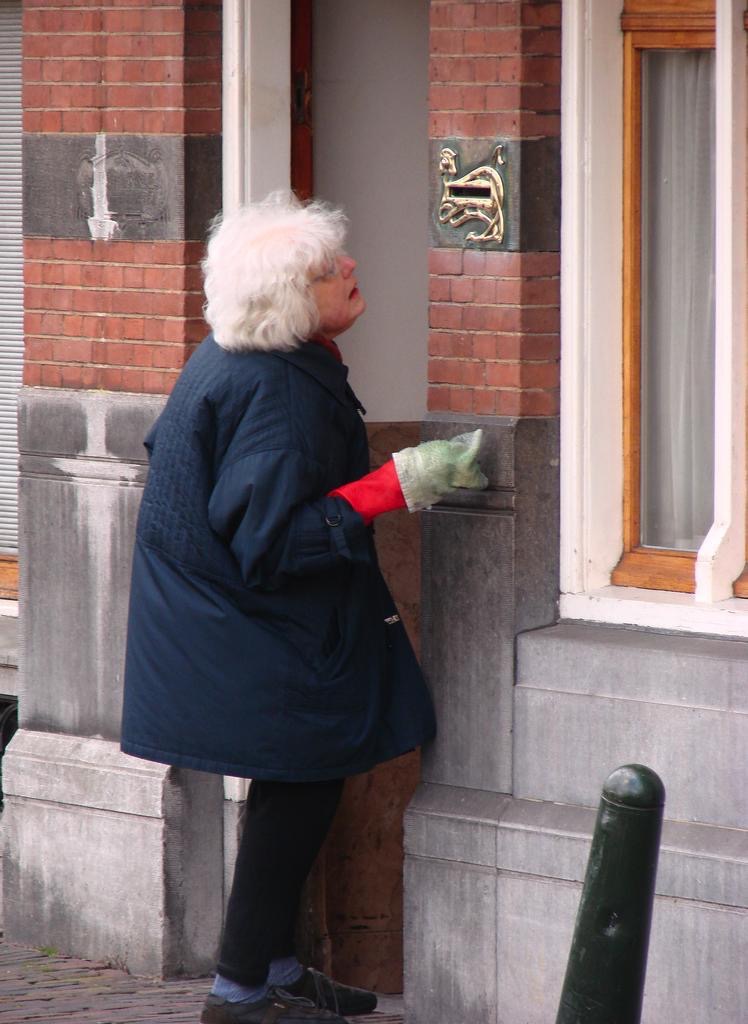} &
		\includegraphics[width=2.4cm,height=3.4cm]{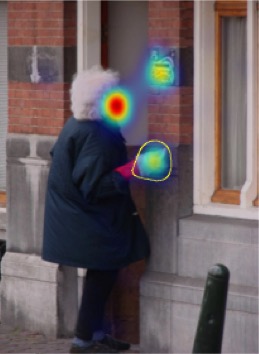} &
		\includegraphics[width=2.4cm,height=3.4cm]{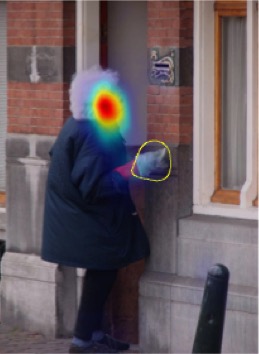} &
		\includegraphics[width=2.4cm,height=3.4cm]{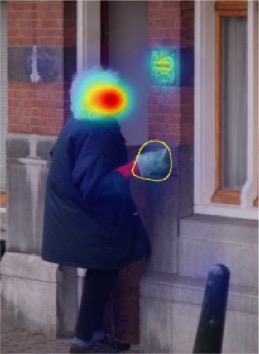} &
		\includegraphics[width=2.4cm,height=3.4cm]{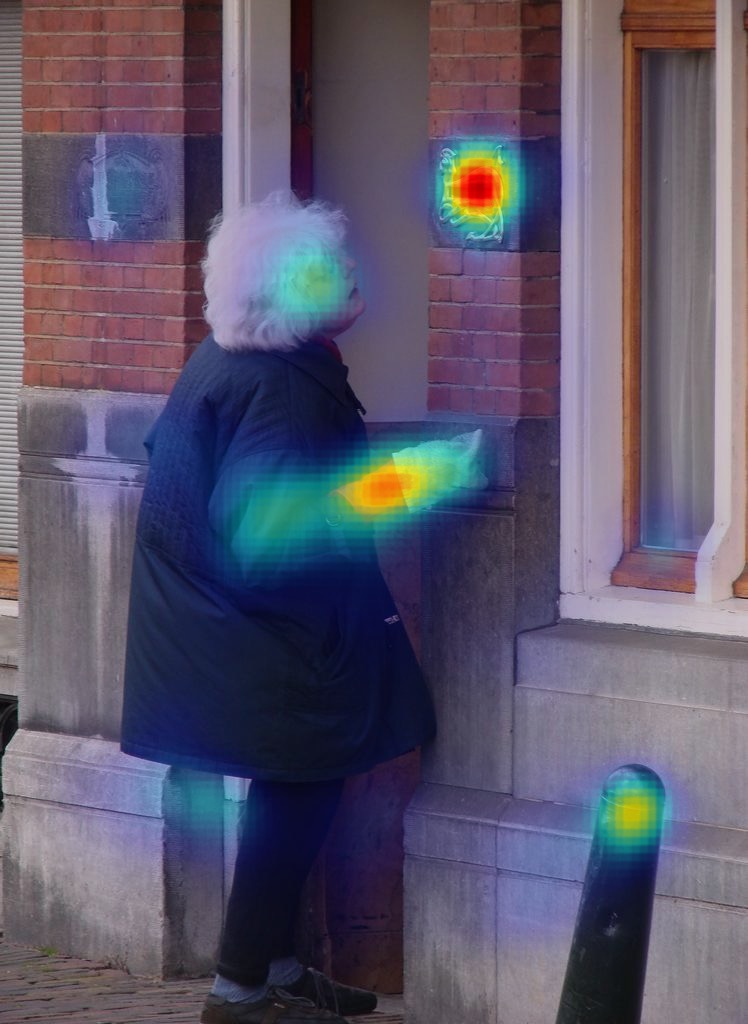} \\
		\includegraphics[width=2.4cm,height=2cm]{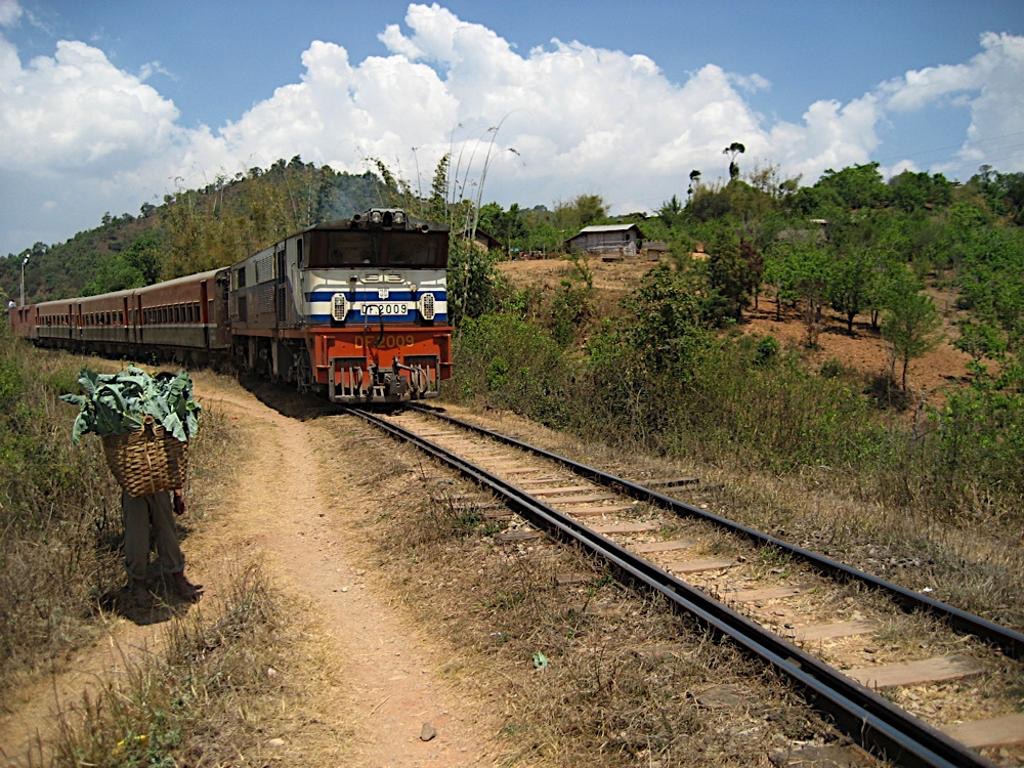} &
		\includegraphics[width=2.4cm,height=2cm]{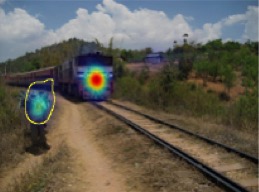} &
		\includegraphics[width=2.4cm,height=2cm]{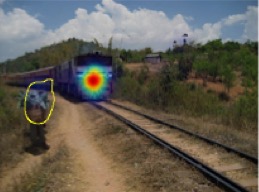} &
		\includegraphics[width=2.4cm,height=2cm]{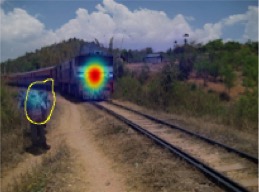} &
		\includegraphics[width=2.4cm,height=2cm]{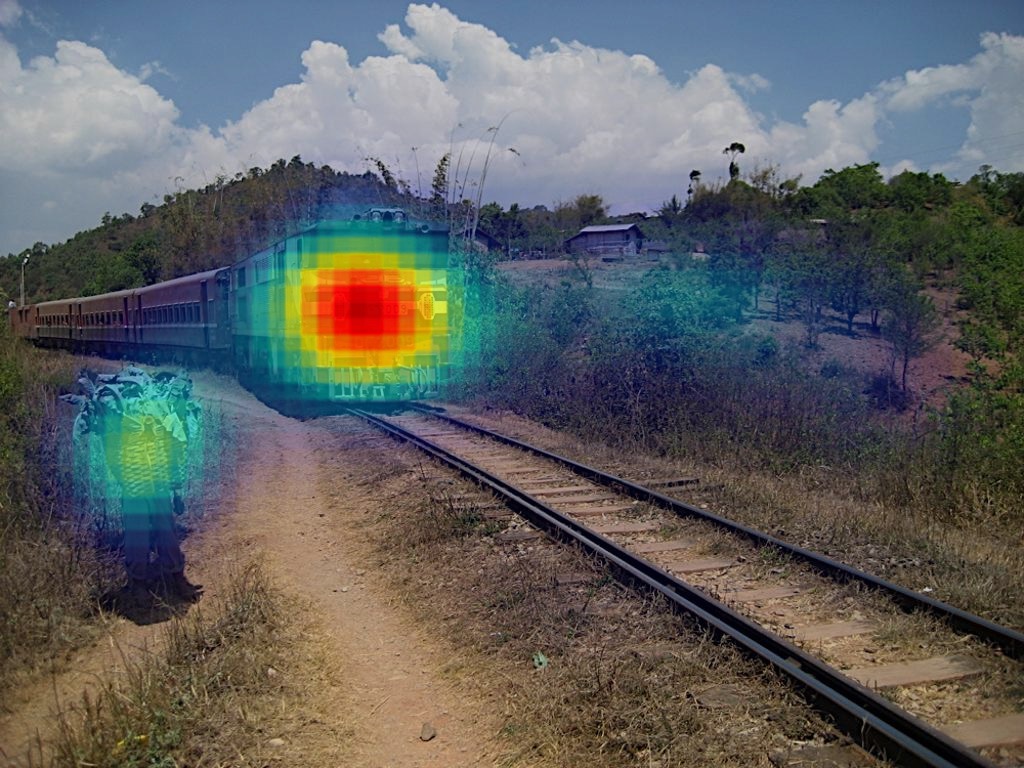} \\
		\includegraphics[width=2.4cm,height=3.5cm]{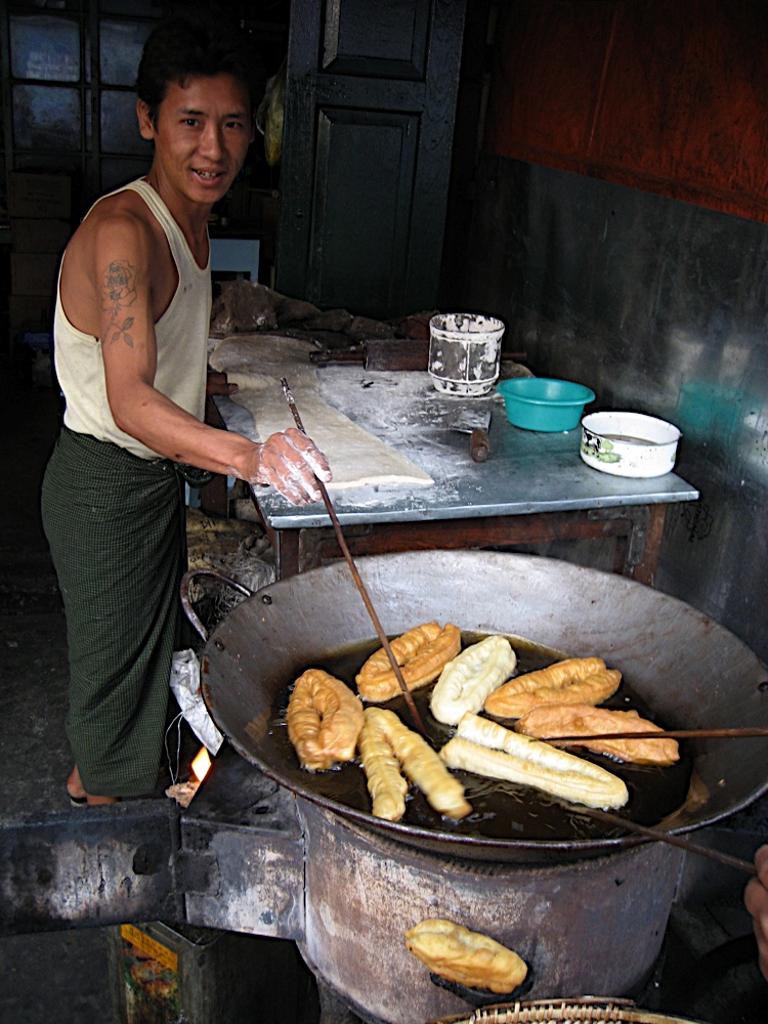} &
		\includegraphics[width=2.4cm,height=3.5cm]{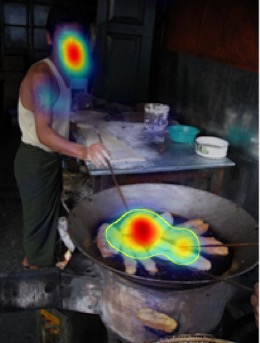} &
		\includegraphics[width=2.4cm,height=3.5cm]{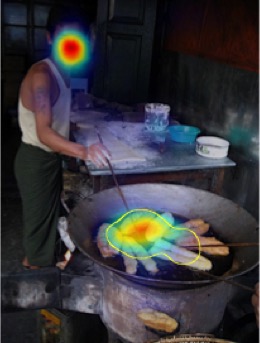} &
		\includegraphics[width=2.4cm,height=3.5cm]{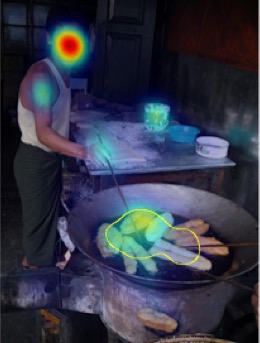} &
		\includegraphics[width=2.4cm,height=3.5cm]{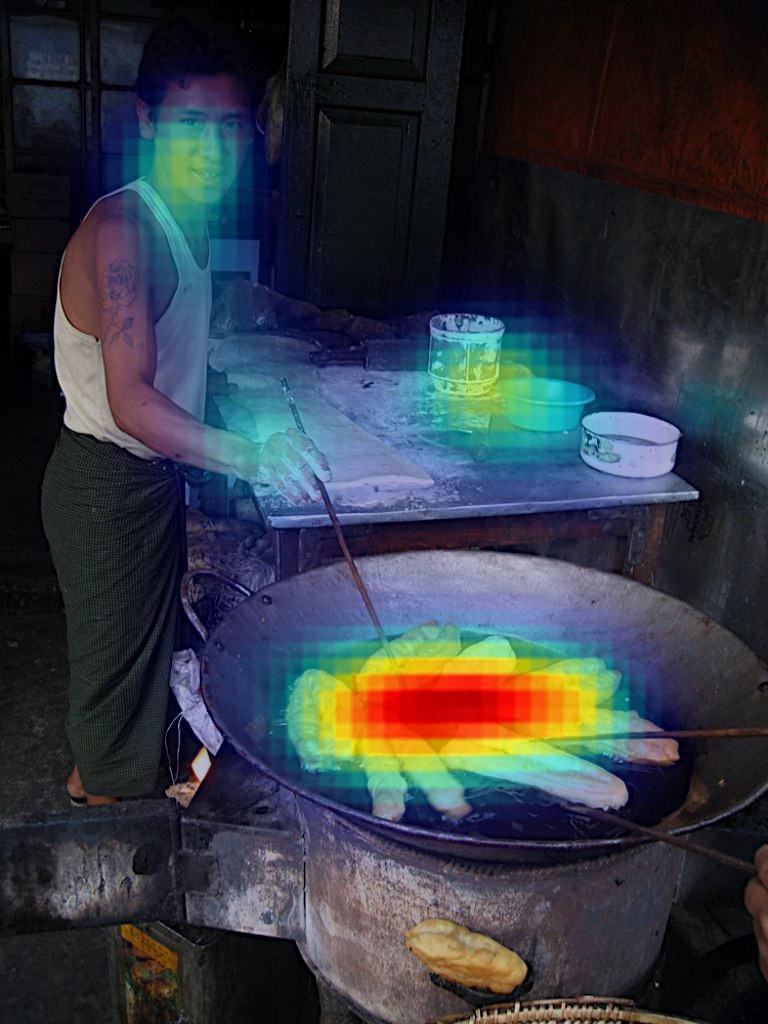} \\
		\includegraphics[width=2.4cm,height=3.2cm]{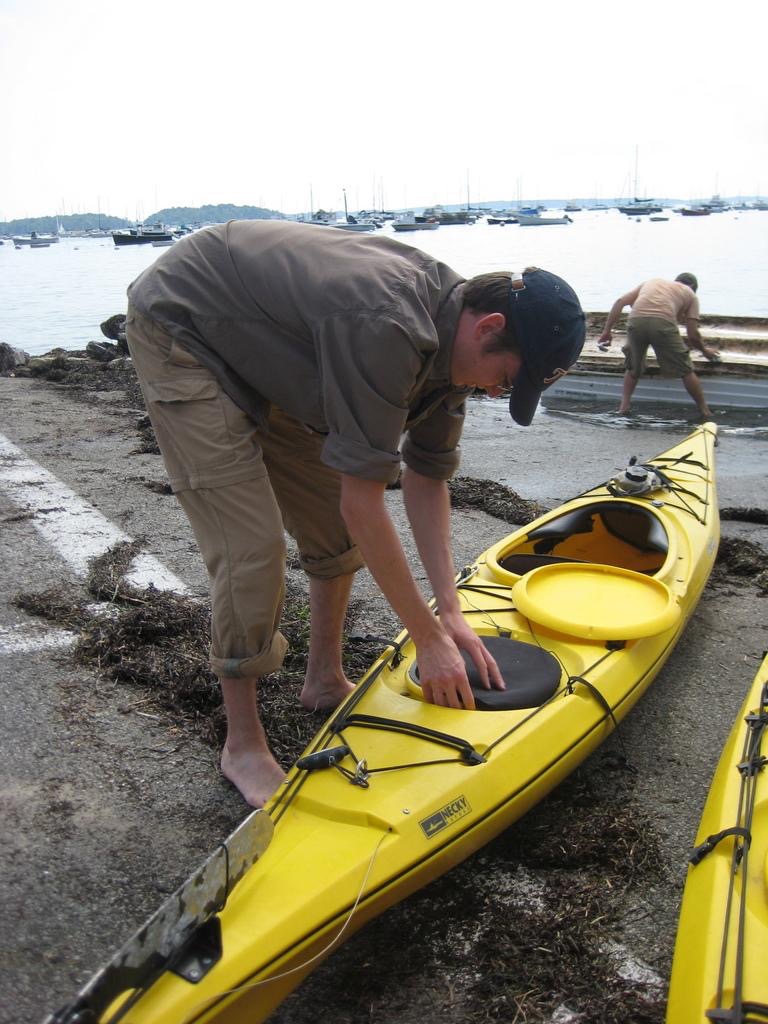} &
		\includegraphics[width=2.4cm,height=3.2cm]{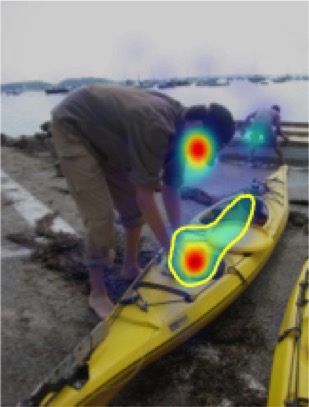} &
		\includegraphics[width=2.4cm,height=3.2cm]{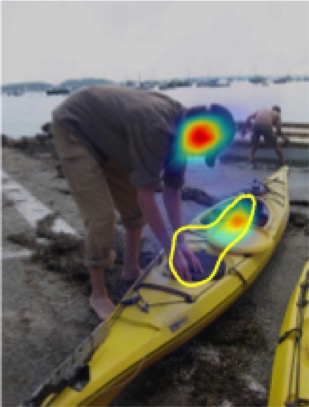} &
		\includegraphics[width=2.4cm,height=3.2cm]{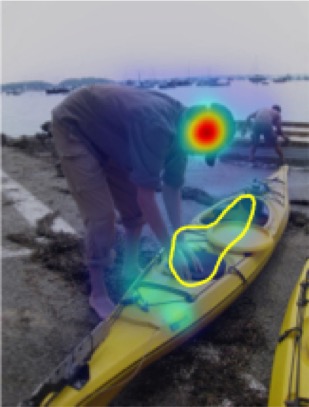} &
		\includegraphics[width=2.4cm,height=3.2cm]{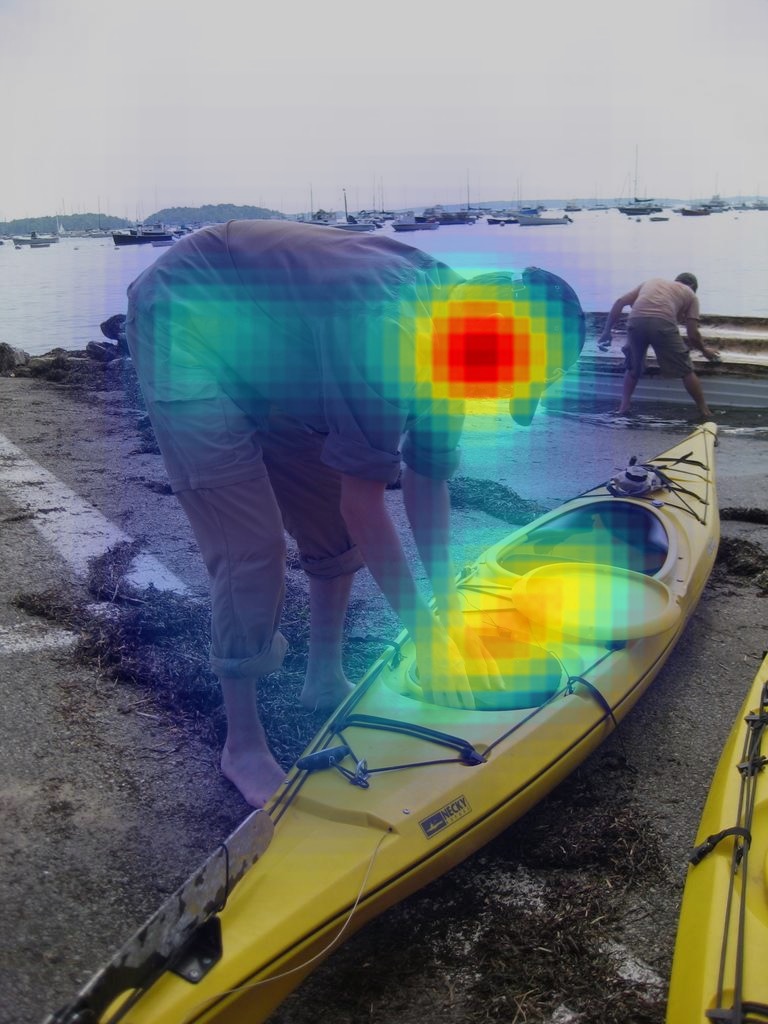} \\
		
		\normalsize Input Image &\normalsize Ground Truth & \normalsize DeepFix~\cite{kruthiventi2015deepfix} & \normalsize SALICON~\cite{jiang2015salicon} & \normalsize STSConvNet
	\end{tabular}}
	\caption{Some experiments on still images. While the top performing deep static saliency models fail to compute satisfactory results in these images (results taken from~\cite{bylinskii2016should}), our spatio-temporal saliency network (STSConvNet) can produce better saliency maps using predicted optical flow maps.}
	\label{fig:motioneffect}
\end{figure}
 
In Table~\ref{tab:ucf-results}, we provide the performance of our model compared to the previous dynamic saliency models which are publicly available on the web. As can be seen, our STSConvNet* model performs better than the state-of-the-art models according to majority of the evaluation measures. It especially outperforms the recently proposed DWS model in terms of all measures. These results suggest that our two-stream network generalizes well beyond the DIEM dataset. 

In Figure~\ref{fig:tab:early-ucf}, we present sample qualitative results of Fang \emph{et al.}~\cite{fang2014} and DWS model~\cite{awsd} (two recently proposed models) and our STSConvNet model on some video frames. For instance, we observe that for the sample frame given in the first row, our model fixates to both the runner and the crowd. For the second and the third sample frames, the compared models do not accurately localize the weight lifter and the cowboys as salient, respectively. Similarly, the proposed STSConvNet* model predicts the eye fixations better than the competing models for the fourth image containing a guardian walking in a corridor. For the last diving image, STSConvNet* and Fang~\emph{et~al.} give results fairly close to the ground truth, while DWS output some spurious regions as salient.

\subsection{Experiments on Still Images from MIT300}

Deep static saliency networks achieve excellent performances on existing benchmark datasets for static saliency estimation. These models, however, only exploit spatial information captured in still images, but sometimes an image, despite being taken in an instant, might carry plenty of information regarding the inherent motion exist in it. In a recent study by Bylinski~\emph{et~al.}~\cite{bylinskii2016should}, the authors demonstrate that the areas showing these kind of activities are indeed evidently important for saliency prediction since humans have tendency to look at the objects that they think in motion or that are in interaction with humans or some other objects. Motivated by these observations, in this section, we present the failures or the shortcomings of the current deep static saliency models through some examples, and show how motion information exist in still images can be utilized to fill in the semantic gap exist in the current static saliency models.

Figure~\ref{fig:motioneffect} presents sample images taken from~\cite{bylinskii2016should} and which are from the MIT 300 dataset~\cite{mit-saliency-benchmark} where highly fixated regions (which cover the 95th percentile of the human fixation maps) are highlighted with yellow curves. As can be clearly seen from these examples, the state-of-the-art deep static models generally fail to give high saliency values to regions where an action occurs or which contains objects that are interpreted as in motion. To capture those regions, we employ the deep optical flow prediction model~\cite{walker2015dense} which extracts optical flow from static images. Once we estimate the motion map of a still image, we can exploit this information together with the RGB image as inputs to our spatio-temporal saliency network (STSConvNet) to extract a saliency map. We observe that using these (possibly noisy) motion maps within our framework provides more accurate predictions than the existing deep static saliency models, and even captures the objects of gaze as illustrated in the first two sample images. These experiments reveal that the performances of static saliency networks can be improved by additionally considering motion information inherent in still images.

\section{Conclusion}
In this work, we have investigated several deep architectures for predicting saliency from dynamic scenes. Two of these	 deep models are single-stream convolutional networks respectively trained for processing spatial and temporal information. Our proposed spatio-temporal saliency networks, on the other hand, are built based on two-stream architecture and employ different fusion strategies, namely direct averaging, max fusion and convolutional fusion, to integrate appearance and motion features, and they are all trainable in an end-to-end manner. While training these saliency networks, we additionally employ an effective and well-founded data augmentation method that utilizes low-resolution versions of the video frames and the ground truth saliency maps, giving a significant boost in performance. Our experimental results demonstrate that the proposed STSConvNet model achieves superior performance over the state-of-the-art methods on DIEM and UCF-Sports datasets. Lastly, we provide some illustrative example results on a number of challenging still images, which show that static saliency estimation can also benefit from motion information. This is left as an interesting topic for future research.

\section*{Acknowledgment}
This research was supported in part by TUBITAK Career Development Award 113E497 and Hacettepe BAP FDS-2016-10202.

\ifCLASSOPTIONcaptionsoff
  \newpage
\fi

\bibliographystyle{IEEEtran}
\bibliography{egbib}

\vspace{-1.5cm}
\begin{IEEEbiography}
	[{\includegraphics[width=1in,height=1.25in,clip,keepaspectratio]{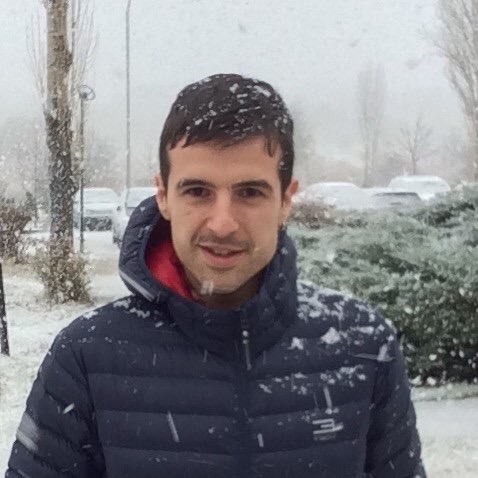}}]\\
	\textbf{Cagdas Bak} received his B.Sc. degree in Computer Engineering from Hacettepe University, Ankara, Turkey in 2013. He got his Master of Science degree from the same deparment in 2016. His current research interests include image and video processing, visual saliency and deep learning.
\end{IEEEbiography}

\begin{IEEEbiography}
	[{\includegraphics[width=1in,height=1.25in,clip,keepaspectratio]{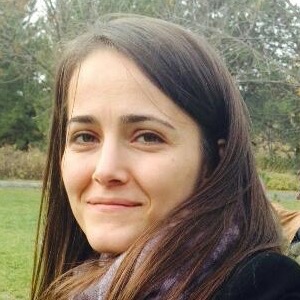}}]\\
	\textbf{Aysun Kocak} received her B.Sc. degree in Computer Engineering from Hacettepe University, Ankara, Turkey in 2012. She is currently pursuing her Ph.D. studies in the same department. Her current research interests include image and video processing, scanpath estimation, visual saliency and deep learning.
\end{IEEEbiography}

\begin{IEEEbiography}[{\includegraphics[width=1in,clip,keepaspectratio]{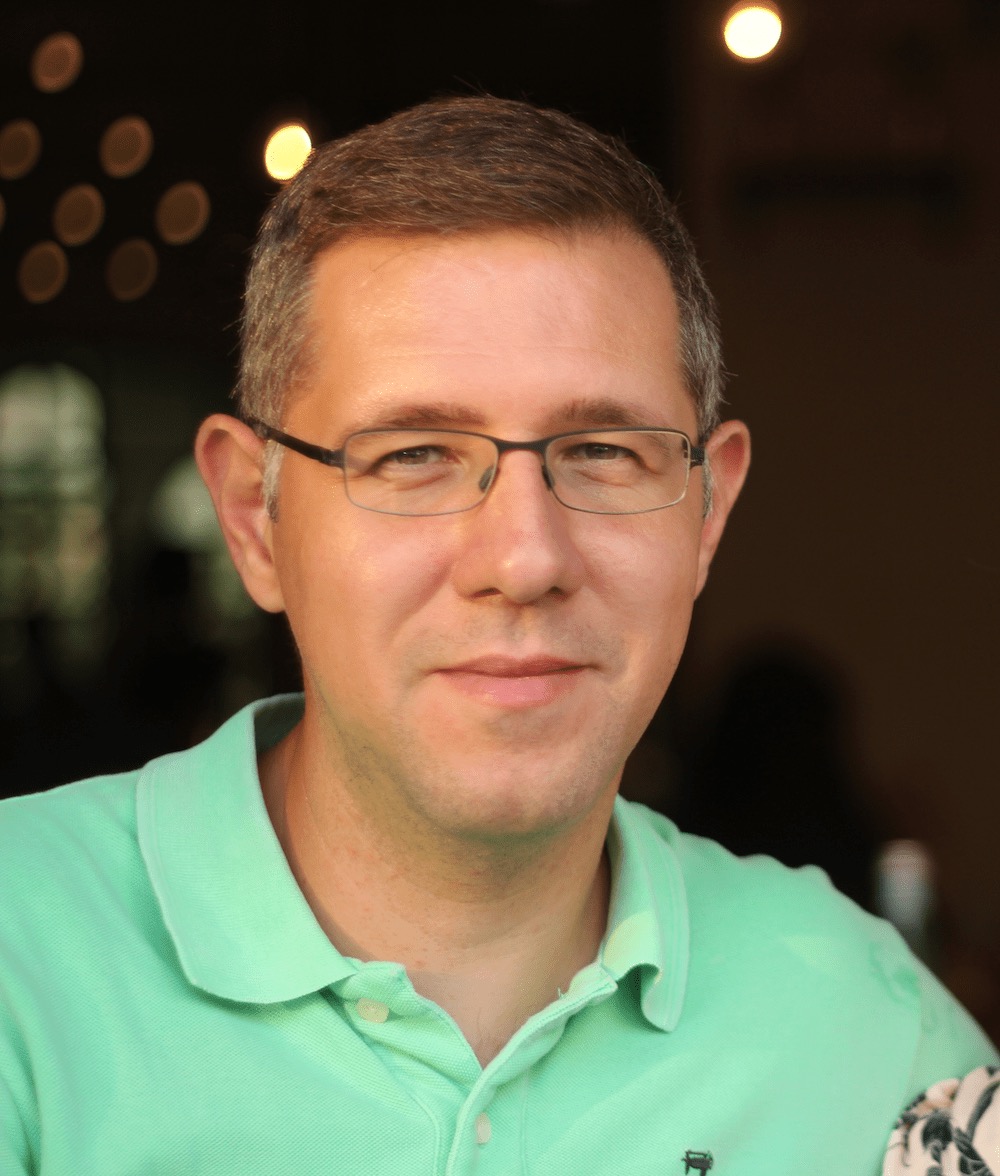}}]
{Erkut Erdem} has received his B.Sc. and M.Sc. degrees respectively in 2001 and 2003 from the Department of Computer Engineering, Middle East Technical University. After completing his Ph.D. work at the Middle East Technical University in 2008, he continued his post-doctoral research studies at T\'{e}l\'{e}com ParisTech, Ecole Nationale Sup\'{e}rieure des T\'{e}l\'{e}communications between 2009 and 2010. He is an Assistant Professor at the Department of Computer Engineering, Hacettepe University since 2014. While his research interests in general concern computer vision and machine learning, he is conducting research activities specifically lie in image editing and smoothing, visual saliency prediction and language and vision.
\end{IEEEbiography}

\begin{IEEEbiography}[{\includegraphics[width=1in,clip,keepaspectratio]{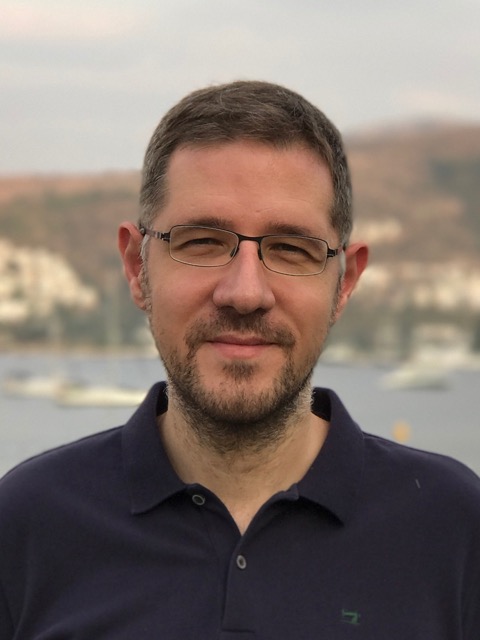}}]
{Aykut Erdem}
	has received his B.Sc. and M.Sc. degrees in Computer Engineering in 2001 and 2003 from Middle East Technical University (METU), Ankara, Turkey. Upon receiving his Ph.D. degree in 2008, he worked as a post-doctoral researcher in the Computer Science Department of Ca’Foscari University of Venice, Italy from 2008-2010. In 2010, he joined Hacettepe University, Ankara, Turkey, where he is now an Assistant Professor at the Department of Computer Engineering. His research interests include computer vision and machine learning, currently focused on image matting, summarization of videos and large image collections, and integrating language and vision.
\end{IEEEbiography}

\end{document}